%% file: manuscript.tex
\newcommand{\changesred}{0}
\newcommand{\red}{\textcolor{red}}
\newcommand{\argmin}{\operatorname*{\text{argmin}}}

\newcommand{\mode}{1}
\newcommand*{\myvspace}{-0.4cm}
\ifnum\mode=0
\newcommand{\CLASSINPUTinnersidemargin}{1.25in}
\newcommand{\CLASSINPUTtoptextmargin}{1.00in}
\newcommand{\CLASSINPUTbottomtextmargin}{1.00in}
\documentclass[draftcls, 11pt, journal, oneside, onecolumn, romanappendices]{IEEEtran}
\else
\documentclass[10pt,letterpaper,journal,oneside,twocolumn]{IEEEtran}
\fi

\usepackage{ifthen}
\usepackage{caption}
\ifthenelse{\equal{\changesred}{1}}{
\captionsetup[table]{name={\color{red}Table}}}
{\captionsetup[table]{name=Table}}

\usepackage{times}
\usepackage{epsfig}
\usepackage{hyperref}
\usepackage{graphicx}
\usepackage{amsmath}
\usepackage{amssymb}
\usepackage{subfig}
\usepackage{color}
\usepackage{tabu}
\usepackage{anyfontsize}
\usepackage[utf8]{inputenc}
\InputIfFileExists{./fig_source/tu_bs_colors}{}{}

\usepackage{multirow}
\usepackage{booktabs}

\begin{document}
	\ifthenelse{\equal{\mode}{0}}
	{
		\ifthenelse{\equal{\changesred}{1}}{
			\title{\red{Vulnerability of Semantic Segmentation\\Networks}  To Adversarial Attacks \red{in} Autonomous Driving}}
		{\title{Vulnerability of Semantic Segmentation\\Networks To Adversarial Attacks in Autonomous Driving}}
	}	
	{
		\ifthenelse{\equal{\changesred}{1}}{
			\title{\red{Vulnerability of Semantic Segmentation Networks}\\To Adversarial Attacks \red{in} Autonomous Driving}}
		{\title{{\huge The Vulnerability of Semantic Segmentation Networks\\to Adversarial Attacks in Autonomous Driving:\\Enhancing Extensive Environment Sensing}}}
	}
	
	\ifthenelse{\equal{\mode}{0}}
	{
		\author{Andreas~Bär$^{1}$,~\IEEEmembership{Student Member,~IEEE,} Jonas~Löhdefink$^{1}$,~\IEEEmembership{Student Member,~IEEE,}\\[0.5em]
			Nikhil~Kapoor$^{2}$, Serin~J.~Varghese$^{2}$, Fabian~Hüger$^{2}$, Peter~Schlicht$^{2}$\\[0.5em]
			and Tim~Fingscheidt$^{1}$,~\IEEEmembership{Senior Member,~IEEE}
			\thanks{$^{1}$Andreas Bär, Jonas Löhdefink and Tim Fingscheidt are with the Institute for Communications Technology,
				Technische Universität Braunschweig, Schleinitzstr. 22, 38106 Braunschweig, Germany
				{\tt\footnotesize \{andreas.baer, j.loehdefink, t.fingscheidt\}@tu-bs.de}}%
			\thanks{$^{2}$Nikhil Kapoor, Serin J.~Varghese, Fabian Hüger and Peter Schlicht are with Volkswagen Group Innovation, Innovation Center Europe, Berliner Ring 2, 38440 Wolfsburg, Germany
				{\tt\footnotesize \{nikhil.kapoor, john.serin.varghese, fabian.hueger, peter.schlicht\}@volkswagen.de}}}
	}
	{
		\author{Andreas~Bär$^{1}$, Jonas~Löhdefink$^{1}$, Nikhil~Kapoor$^{2}$, Serin~J.~Varghese$^{2}$, \\[0.5em] Fabian~Hüger$^{2}$, Peter~Schlicht$^{2}$ and Tim~Fingscheidt$^{1}$,~\IEEEmembership{Senior Member,~IEEE}
			\thanks{$^{1}$Andreas Bär, Jonas Löhdefink and Tim Fingscheidt are with the Institute for Communications Technology,
				Technische Universität Braunschweig, Schleinitzstr. 22, 38106 Braunschweig, Germany
				{\tt\footnotesize \{andreas.baer, j.loehdefink, t.fingscheidt\}@tu-bs.de}}
			\thanks{$^{2}$Nikhil Kapoor, Serin J.~Varghese, Fabian Hüger and Peter Schlicht are with Volkswagen Group Automation, Berliner Ring 2, 38440 Wolfsburg, Germany
				{\tt\footnotesize \{nikhil.kapoor, john.serin.varghese, fabian.hueger, peter.schlicht\}@volkswagen.de}}}
	}

\maketitle
\begin{abstract}
Enabling autonomous driving (AD) can be considered as one of the biggest challenges in today's technology.
AD is a complex task accomplished by several functionalities, with the environment perception being one of its core functions.
Environment perception is usually performed by combining the semantic information captured by several sensors, i.e., light detection and ranging (LiDAR) or camera.
The semantic information from the respective sensor can be extracted by using convolutional neural networks (CNNs) for dense prediction.
In the past years, CNNs constantly showed state-of-the-art performance on several vision-related tasks, such as semantic segmentation of traffic scenes using nothing but RGB images provided by a camera.
Although CNNs obtain state-of-the-art performance on clean images, almost imperceptible changes to the input, referred to as adversarial perturbations, may lead to fatal deception.
The goal of this article is to illuminate \ifthenelse{\equal{\changesred}{1}}{\red{
vulnerability
}}
{vulnerability
}aspects of CNNs used for semantic segmentation with respect to adversarial attacks, and share insights into some of the existing known adversarial defense strategies.
We aim at clarifying the advantages and disadvantages coming with applying CNNs for environment perception in AD in order to serve as a motivation for future research in this field.
\end{abstract}

\section{Introduction}
The desire for mobility is a driving force in progressing technology, with \textit{autonomous driving} (AD) clearly being the next major step in automotive technology along with electromobility.
An AD vehicle is a highly complex system with several sensors and subcomponents, one of them being \textit{vehicle-to-everything} (V2X) communication.

\ifthenelse{\equal{\mode}{0}}{}{
	\begin{figure}[t!]
		\centering
		\includegraphics[width=\columnwidth]{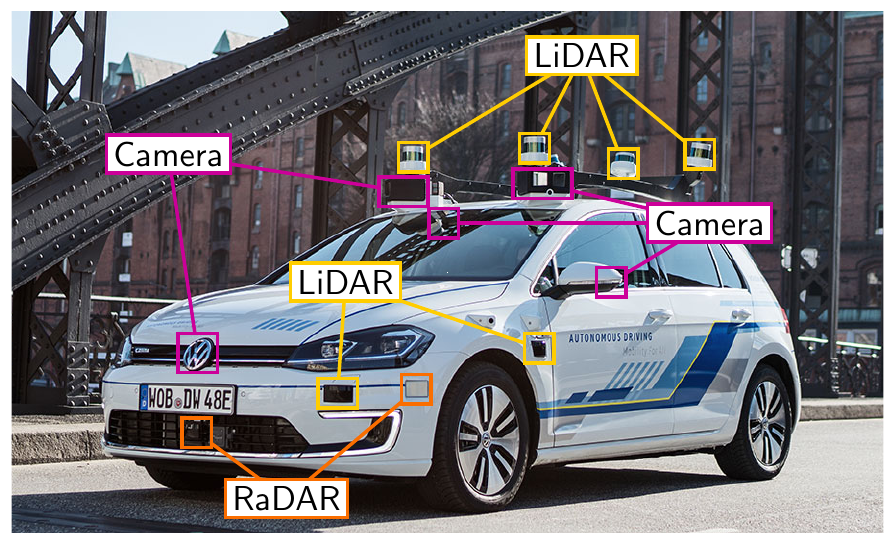}
		\caption[toc entry]{An autonomous driving (AD) research vehicle \ifthenelse{\equal{\changesred}{1}}{\red{
					equipped with
			}}
			{equipped with
			}radio detection and ranging (RaDAR, colored in orange), light detection and ranging (LiDAR, colored in yellow), and camera \ifthenelse{\equal{\changesred}{1}}{\red{
					sensors
			}}
			{sensors
			}(colored in purple). The sensors are placed at different locations to obtain an extensive environment sensing.}\vspace{\myvspace}
		\label{fig:car}
\end{figure}}

In the context of AD, V2X communication has several applications, e.g., path planning \ifthenelse{\equal{\changesred}{1}}{\red{and}}{and} decision making~\cite{Zeng2019}, \ifthenelse{\equal{\changesred}{1}}{\red{or}}{or} systems for localization \ifthenelse{\equal{\changesred}{1}}{\red{and}}{and} cooperative perception~\cite{kim2015impact}.
All autonomous systems need a perception stage which constitutes the first step in the process chain of sensing the environment.
The purpose of cooperative perception systems in AD is the exploitation of information stemming from other traffic participants to increase safety, efficiency and comfort aspects while driving~\cite{hobert2015enhancements}.
The common concept lies in information transmission between various vehicles as well as between vehicles and back-end servers over any kind of (wireless) transmission channel.
The transmitted information ranges from trajectories of the ego vehicle and other traffic participants over vehicle state information to sensor data coming from radio detection and ranging (RaDAR), light detection and ranging (LiDAR), and camera, and assists in constructing a more complete model of the physical world.

Each decision \ifthenelse{\equal{\changesred}{1}}{\red{of}}{of} an AD vehicle is based on the underlying \textit{environment perception} and is intended to lead to an appropriate action.
Hence, the proper perception of the environment is an essential ingredient for reducing road accidents to a bare minimum to foster public acceptance of AD.
The most common sensors of a single AD vehicle's environment perception system (\cite{Bengler2014, Levinson2011, Wei2013}) are illustrated in Fig.~\ref{fig:car}.

\ifthenelse{\equal{\mode}{0}}{}{
\begin{figure*}[t!]
	\centering
	\includegraphics[width=0.245\textwidth, height=0.125\textwidth]
	{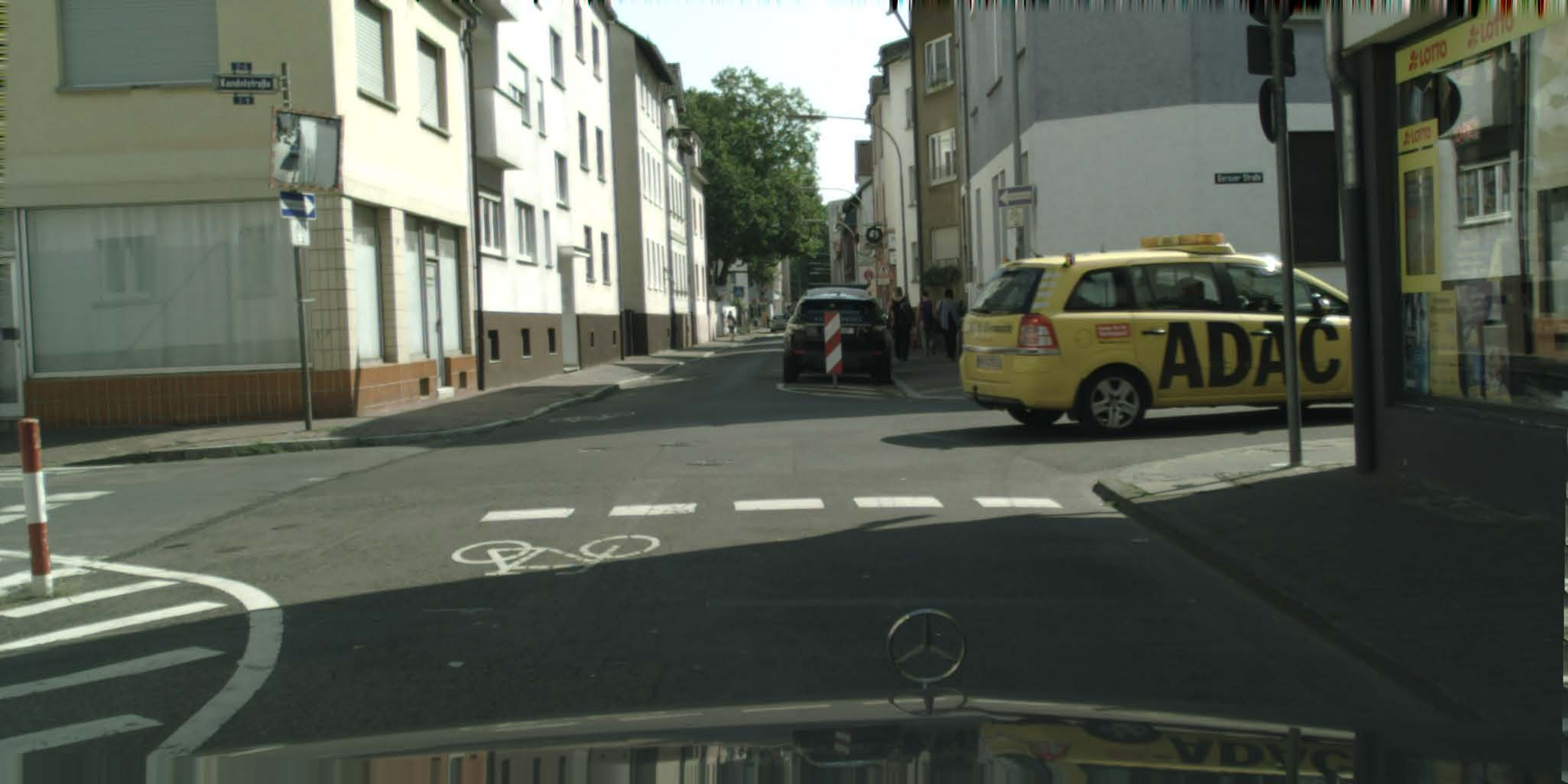}\vspace{0.1cm}
	\includegraphics[width=0.245\textwidth, height=0.125\textwidth]
	{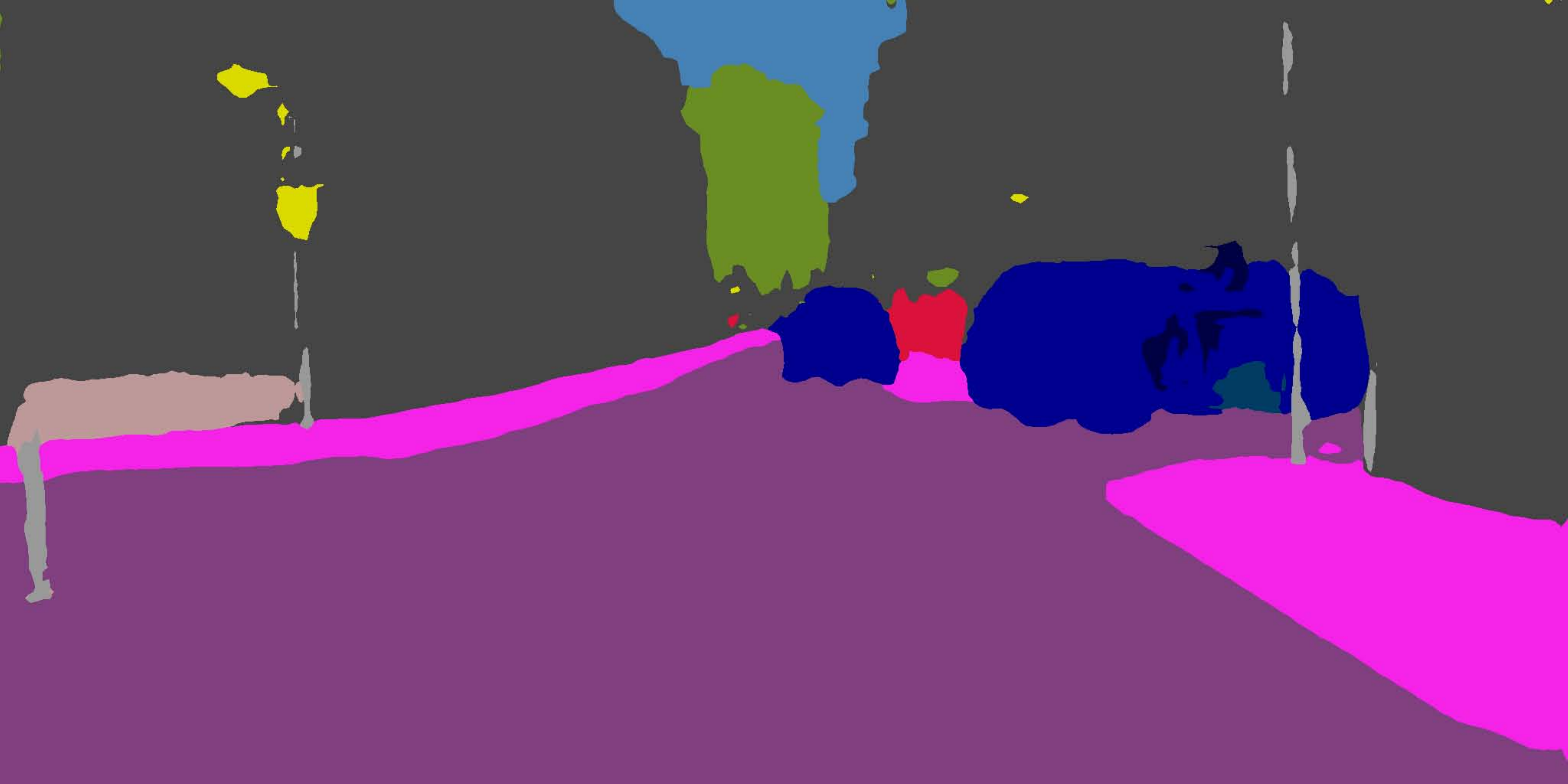}
	\includegraphics[width=0.245\textwidth, height=0.125\textwidth]
	{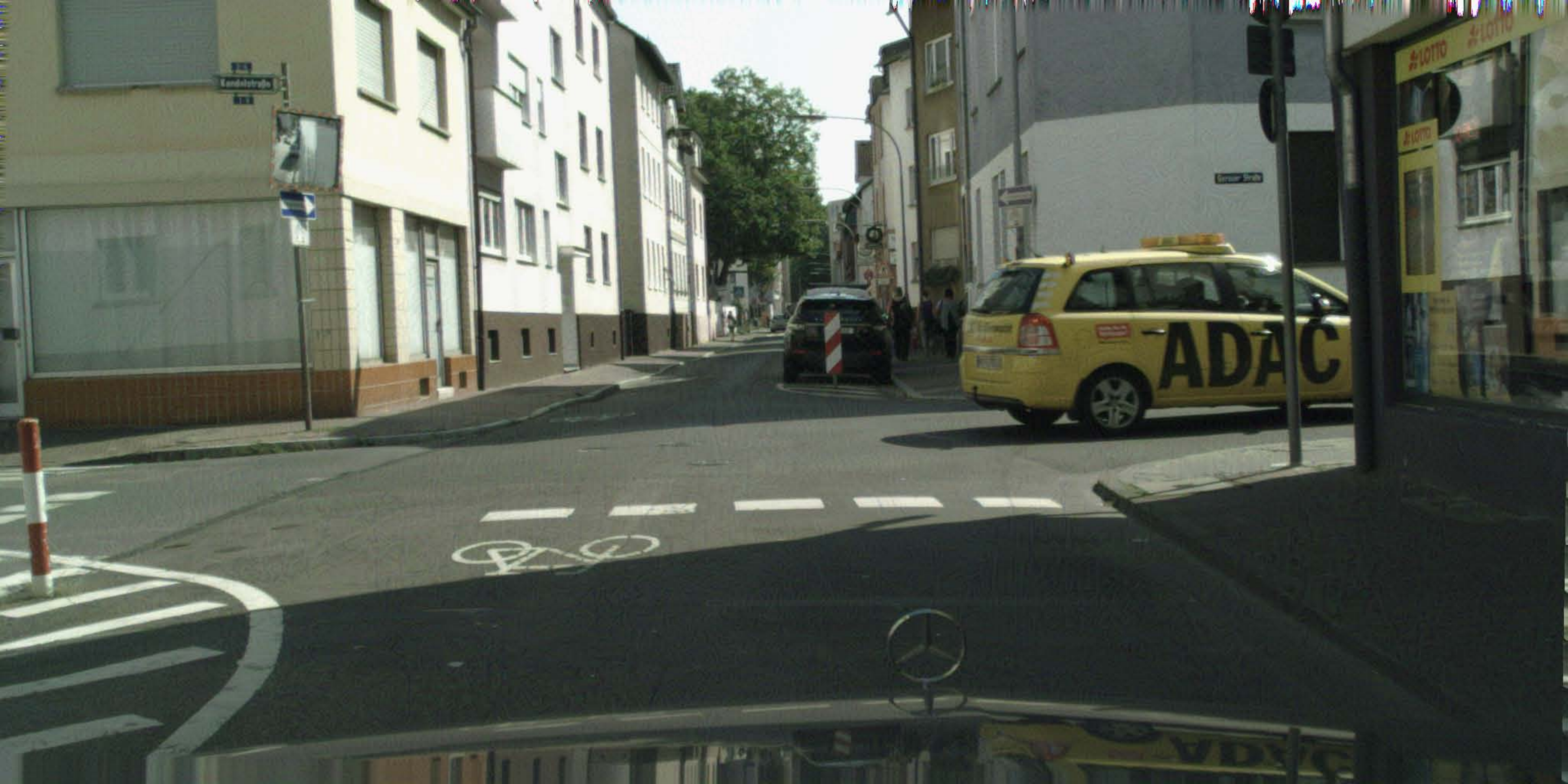}
	\includegraphics[width=0.245\textwidth, height=0.125\textwidth]
	{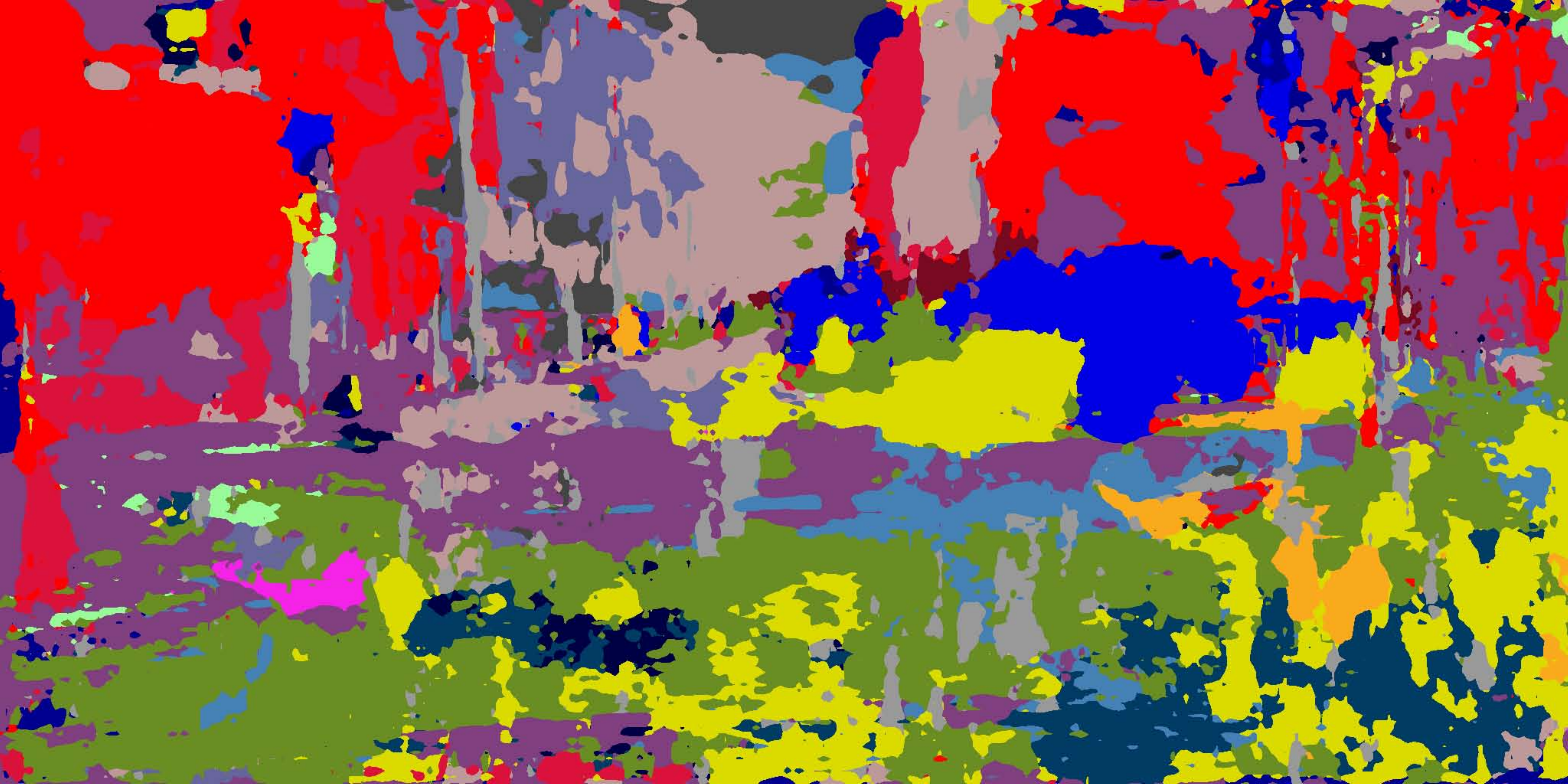}
				
	\begin{tabu} to \textwidth {X[c]X[c]X[c]X[c]}
		(a) & (b) & (c) & (d)
	\end{tabu}
	
	\caption[toc entry]{A simple adversarial attack using the iterative least-likely class method (LLCM)~\cite{Kurakin2017a} to fool the ICNet \cite{Zhao2018a} on \protect\ifthenelse{\equal{\changesred}{1}}{\red{a hand-picked image from}}{a hand-picked image from} the Cityscapes validation set; (a) clean input image, (b) semantic segmentation of clean input image, (c) adversarial example, and (d) semantic segmentation of adversarial example.}\vspace{\myvspace}
	\label{fig:adv_llm_images}
\end{figure*}}

	Several external sensors, i.e., RaDAR, LiDAR, and camera, are mounted on an AD vehicle.
RaDAR sensors are already widely used in multiple automotive functions and are considered to play a key role in enabling AD (\cite{Engels2017, Patole2017}).
LiDAR sensors are capable of detecting obstacles \cite{Levinson2011} and were already used in numerous AD competitions \cite{Bengler2014}.
Camera sensors on the other hand are mainly used for detecting lane markings or traffic signs \cite{Wei2013}, but can also be used for object detection and semantic segmentation \cite{Zhao2018a}.
The data captured by the three sensor groups is gathered within a central processing unit to extract semantic information from the environment.

Over the past few years, the interest \ifthenelse{\equal{\changesred}{1}}{\red{in employing deep neural networks (DNNs)}}{in employing deep neural networks (DNNs)} increased noticeably as they constantly achieved state-of-the-art performance in multiple vision-related tasks and benchmarks, including \textit{semantic segmentation} for AD (\cite{Cordts2016, Long2015}).
Semantic segmentation is a classical computer vision task, where each pixel of an RGB image is assigned to a corresponding semantic class, see Fig.~\ref{fig:adv_llm_images} (a), (b).
Since such camera-based technology is both cheaper and uses less data compared to LiDAR-based technology, it is of special interest for AD. 
Recent progress in semantic segmentation enables real-time processing \cite{Zhao2018a}, making this an even more promising technology for AD applications.

Nevertheless, the environment perception system of an AD vehicle is a highly safety-relevant function.
Any error can lead to catastrophic outcomes in the real world.
While DNNs revealed promising functional performance in a wide variety of tasks, they \ifthenelse{\equal{\changesred}{1}}{\red{show vulnerability}}{show vulnerability} to certain input patterns, denoted as \textit{adversarial examples} \cite{Szegedy2014}.
Adversarial examples are almost imperceptibly \ifthenelse{\equal{\changesred}{1}}{\red{altered}}{altered} versions of an image and are able to fool state-of-the-art DNNs in a highly robust manner, see Fig.~\ref{fig:adv_llm_images} (c), (d).
Assion et al.~\cite{Assion2019} showed that a virtually unlimited set of adversarial examples can be created on each state-of-the-art machine learning model.
This intriguing property of DNNs is of special concern, when looking at their applications in AD \ifthenelse{\equal{\changesred}{1}}{\red{and needs to be addressed further by DNN certification methods (\cite{Dvijotham2018, Wu2018})}}{and needs to be addressed further by DNN certification methods (\cite{Dvijotham2018, Wu2018})} or means of uncertainty quantification \cite{Michelmore2019}.
Cooperative perception for example can be seen as one of the weak spots in the data processing during the environment perception of an AD vehicle.
It can be used as a loophole to intrude adversarial examples to fool AD vehicles in range.
Note, this is only one of many possible scenarios how adversarial examples can find their way into the system.

In this article, we will examine the \ifthenelse{\equal{\changesred}{1}}{\red{vulnerability}}{vulnerability} of DNNs towards adversarial attacks, while focusing on environment perception for AD.
For this purpose we chose semantic segmentation as the underlying function we want to perform adversarial attacks on, since it is a promising technology for camera-based environment perception.
The remainder of this article is structured as follows: First, we give a brief overview of semantic segmentation and introduce the ICNet \cite{Zhao2018a} as a potential network topology, which we will then adopt for our experiments.
Second, we continue with adversarial attacks, starting with simple image classification and extending to adversarial attacks for semantic segmentation.
We demonstrate several visual examples to raise awareness for DNNs' vulnerability towards adversarial attacks.
Third, we examine techniques for defending against the adversarial attacks shown before and compare the obtained qualitative results.
Lastly, we conclude by providing final remarks and discuss some future research directions \ifthenelse{\equal{\changesred}{1}}{\red{pointing out that certification is an important aspect to ensure a certain level of robustness when employing DNNs}}{pointing out that certification is an important aspect to ensure a certain level of robustness when employing DNNs}.
The article is intended to sensibilize the reader towards \ifthenelse{\equal{\changesred}{1}}{\red{vulnerability}}{vulnerability} issues of DNNs in environment perception for AD and to stir interest in the development of new defense strategies for adversarial attacks.

\section{Semantic Segmentation}
An RGB image is a high-dimensional source of data, with pixels being the smallest units of semantic information.
Semantic segmentation is a popular method to extract the semantic information from an RGB image, \ifthenelse{\equal{\changesred}{1}}{\red{where each pixel is tagged}}{where each pixel is tagged} with a label taken from a \ifthenelse{\equal{\changesred}{1}}{\red{finite}}{finite} set of classes.
Today's state of the art in semantic segmentation is dominated by convolutional neural networks (CNNs)\ifthenelse{\equal{\changesred}{1}}{\red{, a special form of DNNs}}{, a special form of DNNs}.
This section introduces some mathematical notation regarding CNNs and gives an overview of the CNN architecture used for semantic segmentation throughout this article.

\subsection{Mathematical Notation}
For the sake of simplicity, we first assume having a CNN, which takes one input image and outputs only a corresponding class for the entire image.
\ifthenelse{\equal{\changesred}{1}}{\red{Hence}}{Hence}, we begin with simple image classification and \ifthenelse{\equal{\changesred}{1}}{\red{then}}{then} extend to semantic segmentation.

\ifthenelse{\equal{\mode}{0}}{}
{
	\begin{figure*}[t!]
		\centering
		\includegraphics[width=\textwidth]{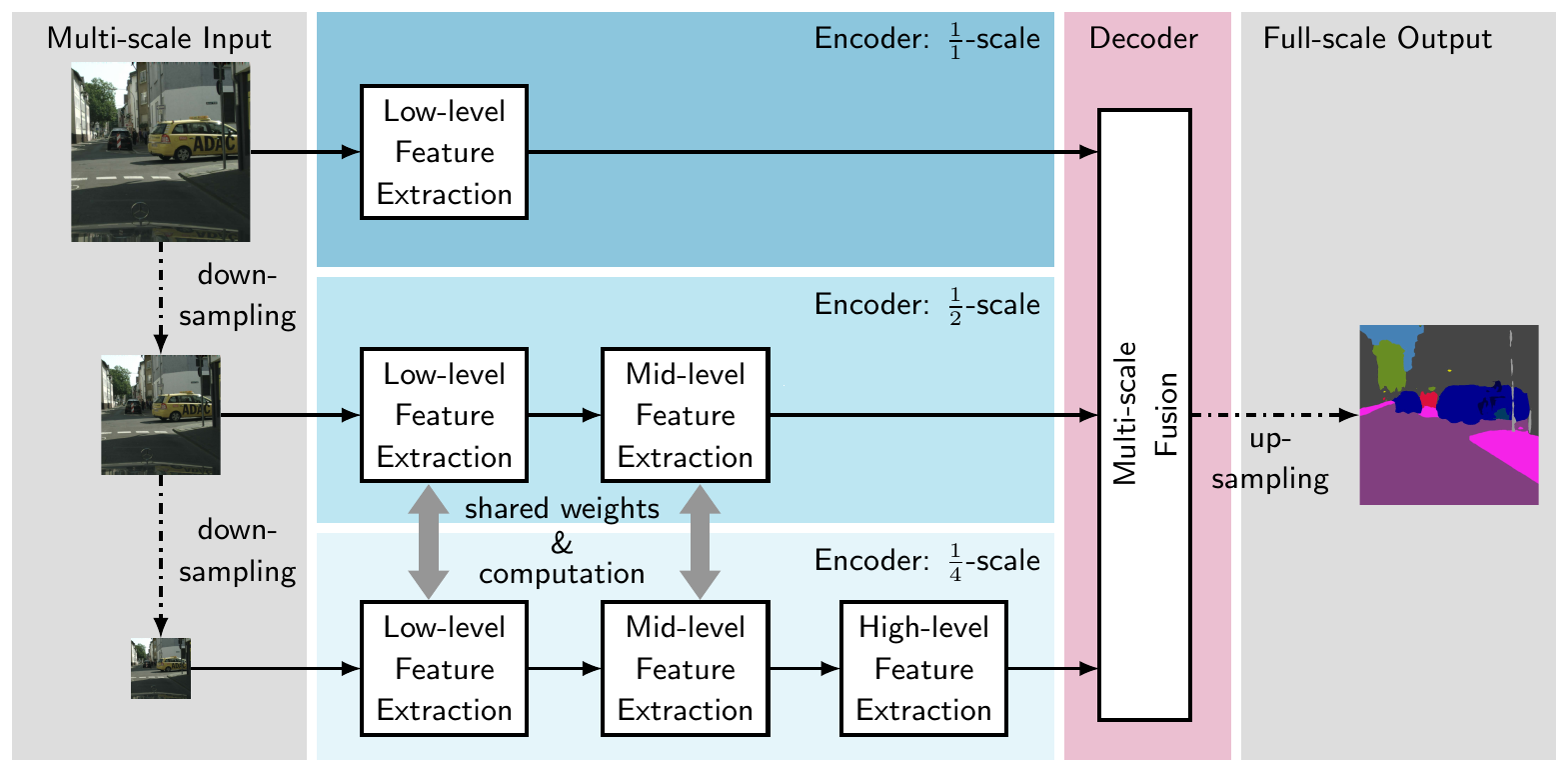}
		\caption{Architectural overview of ICNet~\cite{Zhao2018a}. The ICNet takes different scales of an RGB image as inputs (left gray block) to output a semantic segmentation mask (right gray block). The encoder consists of three scale-dependent parts to extract multi-scale features from the inputs (\textcolor{tu5}{shades of blue}). Each of these three encoder parts perform a downsampling by a factor of eight during feature extraction. To save computational complexity, the bigger scales are limited to low-level and mid-level feature extraction. The extracted multi-scale features are then fused within the decoder by a multi-scale fusion block (\textcolor{tu3}{light magenta}), before performing final upsampling to obtain a full-resolution semantic segmentation mask with respect to the input.}\vspace{\myvspace}
		\label{fig:icnet}
	\end{figure*}
}

First of all, the input image is denoted as $\boldsymbol{x}\in\mathcal{X}\subset\mathbb{G}^{H\times W\times C}$, with image height in pixels $H$, image width in pixels $W$, number of color channels $C$, dataset $\mathcal{X}$, and the set of integer gray values $\mathbb{G}$.
Each image contains gray values $x_i\in\mathbb{G}^{C}$ at each pixel position $i\in\mathcal{I}$, with $\mathcal{I}$ being the set of pixel positions, having the cardinality $|\mathcal{I}|=H\cdot W$.
Smaller patches of an image are denoted as $\boldsymbol{x}_{\mathcal{I}_i} \in\mathbb{G}^{h\times w\times C}$, with patch height in pixels $h$, crop width in pixels $w$, and the set of pixel positions $\mathcal{I}_i \subseteq\mathcal{I}$ with $i$ being the center pixel and $|\mathcal{I}_i|=h\cdot w$.
For the special case of $\mathcal{I}_i=\mathcal{I}$, we obtain $\boldsymbol{x}_{\mathcal{I}_i}  = \boldsymbol{x}$.
A CNN usually consists of several layers $\ell\in\mathcal{L}$ containing feature map activations $\boldsymbol{f}_\ell\!\left( \boldsymbol{x} \right) \in\mathbb{R}^{H_\ell\times W_\ell\times C_\ell}$ of the respective layer $\ell\in\mathcal{L}$, and 1st layer input image $\boldsymbol{x}$, with the set of layers $\mathcal{L}$, feature map height $H_\ell$, feature map width $W_\ell$, and number of feature maps $C_\ell$.
Fed with the input image $\boldsymbol{x}$, a CNN \textit{for image classification} outputs a probability score $P\!\left( s | \boldsymbol{x} \right) \in\mathbb{I}$ for each class $s\in\mathcal{S}$, with $\mathbb{I}=\left[ 0,1\right]$, and the set of classes $\mathcal{S}$, with the number of classes $N=|\mathcal{S}|$, leading to
\begin{equation}
\mathfrak{F}_\text{classification}: \mathbb{G}^{H\times W\times C} \to \mathbb{I}^{N}.
\end{equation}
For better readability, the CNN parameters $\boldsymbol{\theta}$ are omitted in our notation.
The predicted class $s^*\!\left( \boldsymbol{x} \right) \in\mathcal{S}$ for the input image $\boldsymbol{x}$ is then obtained by
\begin{equation}
s^*\!\left( \boldsymbol{x} \right) = \operatorname*{\text{argmax}}_{s\in\mathcal{S}} P\!\left( s| \boldsymbol{x} \right).
\end{equation}

From now on a CNN is considered, which is capable of performing \textit{semantic segmentation}.
The respective CNN outputs a probability $P\!\left(s | i,\boldsymbol{x} \right)$ for each pixel position $i\in\mathcal{I}$ of the input image $\boldsymbol{x}$ and class $s\in\mathcal{S}$.
Altogether, it outputs class scores $\boldsymbol{p} \!\left( \boldsymbol{x} \right)=\mathfrak{F}_\text{segmentation}\!\left( \boldsymbol{x} \right) \in\mathbb{I}^{H\times W\times N}$ for all pixel positions $i\in\mathcal{I}$ and classes $s\in\mathcal{S}$, leading to
\begin{equation}
\mathfrak{F}_\text{segmentation}: \mathbb{G}^{H\times W\times C} \to \mathbb{I}^{H\times W\times N}.
\end{equation}
The semantic segmentation mask $\boldsymbol{m}\!\left( \boldsymbol{x} \right) \in\mathcal{S}^{H\times W}$ containing the predicted class $m_i\!\left( \boldsymbol{x} \right) = s^*_i\!\left( \boldsymbol{x} \right)$ at each pixel position $i\in\mathcal{I}$ of the input image $\boldsymbol{x}$ is then obtained by
\begin{equation}
\boldsymbol{m}\!\left( \boldsymbol{x} \right) = \operatorname*{\text{argmax}}_{s\in\mathcal{S}} \boldsymbol{p} \!\left( \boldsymbol{x} \right).
\end{equation}
The performance of such a CNN is measured by the \textit{mean intersection-over-union} (mIoU)
\begin{equation}
\text{mIoU} = \frac{1}{N} \sum_{s\in\mathcal{S}} \frac{\text{TP}\!\left( s \right)}{\text{TP}\!\left( s \right) + \text{FP}\!\left(s \right) + \text{FN}\! \left( s \right)},
\end{equation}
with the class-specific true positives $\text{TP}\!\left( s \right)$, false positives $\text{FP}\!\left( s \right)$, and false negatives $\text{FN}\!\left( s \right)$.

\subsection{Architecture for Semantic Segmentation}
Today's state-of-the-art CNN architectures for semantic segmentation are often based on the work of Long et al.~\cite{Long2015}.
They proposed to use a CNN, pretrained on image classification, as a feature extractor and further extend it to recover the original image resolution.
The extended part is often referred to as the decoder and fulfills the task of gathering, reforming and rescaling the extracted features for the task of semantic segmentation.
One characteristic of this proposed network architecture is the absence of fully connected layers.
Such CNNs are therefore called fully convolutional networks (FCNs). 

Especially for AD, a real-time capable state-of-the-art CNN being robust to minimal changes in the input is needed.
Arnab et al.~\cite{Arnab2018} analyzed the robustness of various CNNs for semantic segmentation towards simple adversarial attacks (\cite{Goodfellow2015, Kurakin2017a}), and concluded that CNNs using the same input with different scales are often most robust. 
\ifthenelse{\equal{\changesred}{1}}{\red{
		The ICNet developed by Zhao et al.~\cite{Zhao2018a} comprises both, a light-weight CNN architecture with multi-scale inputs.
}}
{The ICNet developed by Zhao et al.~\cite{Zhao2018a} comprises both, a light-weight CNN architecture with multi-scale inputs.
}
The overall structure of the ICNet is depicted in Fig.~\ref{fig:icnet}.
The ICNet is designed to extract multi-scale features by taking different scales of the image as inputs.
The extracted multi-scale features are fused before being upsampled to obtain a full-resolution semantic segmentation mask.
The ICNet mainly profits from the combination of high-resolution low-level features (i.e., edges) with low-resolution high-level features (i.e., spatial context).
For the sake of reproducibility, an openly available reimplementation\footnote{https://github.com/hellochick/ICNet-tensorflow} of the ICNet based on TensorFlow is used and tested on the widely applied Cityscapes dataset \cite{Cordts2016}.
Cityscapes serves as a good dataset for exploring CNNs using semantic segmentation for AD, having pixel-wise annotations for 5000 images (validation, training and test set combined), with relevant classes such as pedestrians and cars. 
The reimplementation of the ICNet achieves 67.26\,\% mIoU on the Cityscapes validation set and runs at about 19 fps on our \texttt{Nvidia Tesla P100} and about 26 fps on our \texttt{Nvidia Geforce GTX 1080Ti} with an input resolution of $1024\times 2048$.
These numbers are promising and indicate that semantic segmentation could serve as a technology for the environment perception system of AD vehicles.

\section{Adversarial Attacks}\label{sec:adversarial_attacks}
Although CNNs exhibit state-of-the-art performance in several vision-related fields of research, Szegedy et al.~\cite{Szegedy2014} revealed their \ifthenelse{\equal{\changesred}{1}}{\red{vulnerability}}{vulnerability} towards certain input patterns.
The CNN topologies they investigated were fooled by just adding small and imperceptible patterns to the input image.
An algorithm producing such \textit{adversarial perturbations} is called an \textit{adversarial attack} and a perturbed image is referred to as an \textit{adversarial example}.

Based on the obvervations of Szegedy et al., new approaches arised for crafting adversarial examples more efficiently (\cite{Athalye2018, Carlini2017, Goodfellow2015, Kurakin2017a, Moosavi-Dezfooli2016}) and were even extended to dense prediction tasks, e.g., semantic segmentation (\cite{Assion2019, Metzen2017, Mopuri2018}).
In the following, two types of adversarial attacks will be introduced: \textit{individual adversarial attacks}, aiming at fooling on the basis of one particular input image, as well as \textit{universal adversarial attacks}, aiming at fooling on the basis of a whole bunch of images at the same time.

\subsection{Individual Adversarial Perturbations}
For the sake of simplicity, CNNs \textit{for image classification} are considered in the following to describe the basic nature of targeted and non-targeted adversarial attacks using individual adversarial perturbations.
As shown before, image classification can be easily extended to semantic segmentation.

Common adversarial attacks aim at fooling a CNN, so that the predicted class $s^*\!\left(\boldsymbol{x}\right)$ does not match with the ground truth class $s\!\left(\boldsymbol{x}\right)\in\mathcal{S}$ of the input image $\boldsymbol{x}$.
One example for such type of an adversarial attack is the fast gradient sign method (FGSM) introduced by Goodfellow et al.~\cite{Goodfellow2015}.
FGSM adopts the loss function $J\!\left(s^*\!\left(\boldsymbol{x}\right), s\!\left(\boldsymbol{x}\right)\right)$ that is used during training of the underlying CNN and computes the adversarial examples by
\begin{equation}\label{fgsm}
\begin{split}
\boldsymbol{x}^\text{adv} = \boldsymbol{x} + \boldsymbol{r} = \boldsymbol{x} + \lambda \, \text{sign}\!\left(\nabla_{\boldsymbol{x}}\, J\!\left(s^*\!\left(\boldsymbol{x}\right), s\!\left(\boldsymbol{x}\right)\right) \right),
\end{split}
\end{equation}
with the adversarial perturbation $\boldsymbol{r}\in\mathbb{R}^{H\times W\times C}$, the step size $\lambda\in\mathbb{R}^+$, and the gradient with respect to the input image $\nabla_{\boldsymbol{x}}\, J\left(s^*\!\left(\boldsymbol{x}\right), s\left(\boldsymbol{x}\right)\right)$.
Note that $\text{sign}\!\left(\, \cdot\,\right)\in\lbrace \pm 1 \rbrace^{H\times W\times C}$.
FGSM lets the perturbation $\boldsymbol{r}$ effectively \textit{in}crease the loss in each dimension by manipulating the input image into positive (``+'') gradient direction.
\ifthenelse{\equal{\changesred}{1}}{\red{Thus, one is not limited to use the ground truth $s\!\left(\boldsymbol{x}\right)$ as depicted in (6), but can in fact use the output of the respective DNN $s^*\!\left(\boldsymbol{x}\right)$.}}{Thus, one is not limited to use the ground truth $s\!\left(\boldsymbol{x}\right)$ as depicted in (6), but can in fact use the output of the respective DNN $s^*\!\left(\boldsymbol{x}\right)$.}

Kurakin et al.~\cite{Kurakin2017a} extended FGSM by an iterative algorithm, changing the adversarial perturbation slightly in each iteration by a small $\lambda$.
To prevent the adversarial perturbation's magnitude from getting too large, it is upper-bounded by
\begin{equation}
||\boldsymbol{r}||_\infty \le \epsilon,
\end{equation}
with $\epsilon\in\mathbb{R}^+$ being the upper bound of the infinity norm and $\epsilon \ge \lambda$.
This way, the perceptibility of the adversarial perturbation is controlled by adjusting $\epsilon$ accordingly.
For the iterative case, (\ref{fgsm}) extends to
\begin{equation}
\begin{split}
\boldsymbol{x}^\text{adv}_0 & = \boldsymbol{x},\\
\boldsymbol{x}^\text{adv}_{\tau+1} & = \boldsymbol{x}^\text{adv}_{\tau} + \boldsymbol{r}_{\tau+1}\\
& = \boldsymbol{x}^\text{adv}_{\tau} + \lambda \, \text{sign}\!\left( \nabla_{\boldsymbol{x}} J\!\left( s^*\!\left(\boldsymbol{x}^\text{adv}_{\tau}\right), s\!\left(\boldsymbol{x} \right) \right) \right),
\end{split}
\end{equation}
with $\tau\in\lbrace 0,1,2,... \rbrace$ being the current iteration index and therefore $\boldsymbol{x}^\text{adv}_{\tau}$ the adversarial example at iteration\footnote{The total number of iterations is set by flooring $\lfloor \text{min}\left(\epsilon + 4, 1.25\epsilon\right)\rfloor$.} $\tau$.

Considering AD vehicles, there exists no ground truth for the data being inferred.
\ifthenelse{\equal{\changesred}{1}}{\red{As already pointed out,}}{As already pointed out,} a naive attacking idea in this setup would be finding an adversarial perturbation $\boldsymbol{r}$, such that (classification!)
\begin{equation}
s^*\!\left( \boldsymbol{x^\text{adv}} \right) \ne s^*\!\left( \boldsymbol{x} \right).
\end{equation}

Such an attack is the least-likely class method (LLCM) introduced by Kurakin et al.~\cite{Kurakin2017a}.
LLCM aims at finding an adversarial pertubation $\boldsymbol{r}$ to obtain
\begin{equation}
s^\text{o}\!\left( \boldsymbol{x} \right) = \operatorname*{\text{argmax}}_{s\in\mathcal{S}} P\!\left( s | \boldsymbol{x}^\text{adv} \right) = \argmin_{s\in\mathcal{S}} P\!\left( s | \boldsymbol{x} \right),
\end{equation}
with the least-likely class $s^\text{o}\!\left( \boldsymbol{x} \right)$ of the input image $\boldsymbol{x}$.
Different from before, the adversarial example using LLCM is obtained by taking a step into the negative direction of the gradient with respect to the input image $\boldsymbol{x}$, according to
\begin{equation}
\boldsymbol{x}^\text{adv} = \boldsymbol{x} - \lambda \, \text{sign}\!\left( \nabla_{\boldsymbol{x}} J\!\left( s^*\!\left(\boldsymbol{x}\right), s^\text{o}\!\left( \boldsymbol{x} \right) \right) \right),
\end{equation}
\textit{minimizing} the loss function.
Similar to FGSM, LLCM can also be performed in an iterative fashion, where in each step a small adversarial perturbation is added to the respective input image.

Another well-known approach for crafting adversarial examples is DeepFool \cite{Moosavi-Dezfooli2016} introduced by Moosavi-Dezfooli and his colleagues.
Compared with FGSM and LLCM, DeepFool does not only search for individual adversarial perturbations, but also tries to find the minimal adversarial perturbation, with respect to an $l_p$-norm, changing the network's output.
This leads us to the following equation
\begin{equation}
\boldsymbol{r}_\text{min} = \argmin_{\boldsymbol{r}} ||\boldsymbol{r}||_p, \, \text{s.t.} \,\,\, s^*\!\left(\boldsymbol{x} + \boldsymbol{r}_\text{min}\right) \ne s^*\!\left( \boldsymbol{x} \right),
\end{equation}
with $|| \cdot ||_p$ being the $l_p$-norm restricting the magnitude of $\boldsymbol{r}_\text{min}$.
Moosavi-Dezfooli et al.~primarily experimented with $p=2$, \ifthenelse{\equal{\changesred}{1}}{\red{
		showing DeepFool's superiority in terms of speed and magnitude}}{showing DeepFool's superiority in terms of speed and magnitude} compared to FGSM, when targeting the same error rate for the respective CNN.
We will not go further into detail here, but we refer the interested reader to \cite{Moosavi-Dezfooli2016} for more information about DeepFool instead.

\ifthenelse{\equal{\changesred}{1}}{\red{
		Carlini and Wagner proposed an approach, which showed to be extremely effective regarding adversarial example detection mechanisms \cite{Carlini2017}.
		They use
		\begin{equation}
		\begin{split}
		\boldsymbol{r}_\text{min} = \argmin_{\boldsymbol{r}} \left( ||\boldsymbol{r}||_2 + c \cdot f(\boldsymbol{x} + \boldsymbol{r}) \right), \\
		\text{s.t.} \,\,\, s^*\!\left(\boldsymbol{x} + \boldsymbol{r}_\text{min}\right) \ne s^*\!\left( \boldsymbol{x} \right),
		\end{split}
		\end{equation}
		as an objective function, with $c$ being a hyperparameter and $f(\boldsymbol{x} + \boldsymbol{r})$ being a loss function.
		Athalye et al.~\cite{Athalye2018} adopted this approach and as a result managed to circumvent several state-of-the-art defense mechanisms.
		We refer the interested reader to \cite{Athalye2018} and \cite{Carlini2017} for more fine-grained information about both approaches and their specific variations.
}}
{Carlini and Wagner proposed an approach, which showed to be extremely effective regarding adversarial example detection mechanisms \cite{Carlini2017}.
	They use
	\begin{equation}
	\centering
	\begin{split}
	\boldsymbol{r}_\text{min} = \argmin_{\boldsymbol{r}} \left( ||\boldsymbol{r}||_2 + c \cdot f(\boldsymbol{x} + \boldsymbol{r}) \right), \\
	\text{s.t.} \,\,\, s^*\!\left(\boldsymbol{x} + \boldsymbol{r}_\text{min}\right) \ne s^*\!\left( \boldsymbol{x} \right),
	\end{split}
	\end{equation}
	as an objective function, with $c$ being a hyperparameter and $f(\boldsymbol{x} + \boldsymbol{r})$ being a loss function.
	Athalye et al.~\cite{Athalye2018} adopted this approach and as a result managed to circumvent several state-of-the-art defense mechanisms.
	We refer the interested reader to \cite{Athalye2018} and \cite{Carlini2017} for more fine-grained information about both approaches and their specific variations.
}

So far, we \ifthenelse{\equal{\changesred}{1}}{\red{introduced}}{introduced} adversarial attacks that were successfully applied on image classification.
Arnab et al.~\cite{Arnab2018} did the first extensive analysis on the behavior of different CNN architectures \textit{for semantic segmentation} using FGSM and LLCM, both performed iteratively and non-iteratively.
They report results on a large variety of CNN architectures, including both lightweight CNN architectures and heavyweight CNN architectures.
The main observation was that network models using residual connections are often more robust when it comes to adversarial attacks.
In addition, lightweight CNN architectures tend to be almost equally robust as heavyweight CNN architectures.
In summary, the results on the Cityscapes dataset demonstrated the vulnerability of CNNs in general.
We show a \ifthenelse{\equal{\changesred}{1}}{\red{typical attack}}{typical attack} in Fig.~\ref{fig:adv_llm_images} using the iterative LLCM on the ICNet with the hyper parameters $\lambda=1$ and $\epsilon=8$.
Despite being mostly imperceptible for the human eye, the adversarial example leads to a dramatically altered network output.
To show the overall effect on the Cityscapes validation set, we computed the mIoU ratio for the iterative LLCM and the non-iterative LLCM using different values for $\epsilon$.
The mIoU ratio $Q$ is defined by 
\begin{equation}
Q = \frac{\text{mIoU}_\text{adv}}{\text{mIoU}_\text{clean}},
\end{equation}
with $\text{mIoU}_\text{adv}$ being the mIoU on adversarially perturbed images $\boldsymbol{x}^\text{adv}$, and $\text{mIoU}_\text{clean}$ being the mIoU on clean images $\boldsymbol{x}$.
The results are plotted in Fig.~\ref{fig:adv_llm_curves}.
As expected, the stronger the adversarial perturbation (in terms of $\epsilon$) the lower the mIoU on adversarial examples $\text{mIoU}_\text{adv}$ and thus a lower mIoU ratio $Q$ is obtained.
As pointed out by Arnab et al.~\cite{Arnab2018}, we also observe that the non-iterative LLCM is even stronger than its iterative counterpart, \ifthenelse{\equal{\changesred}{1}}{\red{which contradicts the original observation made by Kurakin et al.~\cite{Kurakin2017a} on image classification.
		Arnab et al.~argue that this phenomenon might be a dataset property of Cityscapes, since the effect does not occur on their second dataset (Pascal VOC 2012).
		Nonetheless, we do not investigate this further as we anyway want to focus on more realistically looking adversarial attacks in the following.
}}
{which contradicts the original observation made by Kurakin et al.~\cite{Kurakin2017a} on image classification.
	Arnab et al.~argue that this phenomenon might be a dataset property of Cityscapes, since the effect does not occur on their second dataset (Pascal VOC 2012).
	Nonetheless, we do not investigate this further as we anyway want to focus on more realistically looking adversarial attacks in the following.
}

Metzen et al.~\cite{Metzen2017} introduced new adversarial attacks for semantic segmentation.
Instead of only fooling the CNN, they additionally wanted the respective CNN to output more realistically looking semantic segmentation masks.
To do so, Metzen et al.~developed two methods.
The first method uses a fake semantic segmentation mask $\boldsymbol{m}\!\left(\boldsymbol{z}\right)$ instead of the original semantic segmentation mask $\boldsymbol{m}\!\left( \boldsymbol{x} \right)$, with $\boldsymbol{x}\ne\boldsymbol{z}\in\mathcal{X}$, meaning that the fake segmentation mask refers to an existing image of the dataset $\mathcal{X}$.
The overall assumption of Metzen et al.~is that a possible attacker might invest time to create a few uncorrelated fake semantic segmentation masks himself.
Assuming that the attacker wants to use the same fake semantic segmentation mask to fool the respective CNN on several images, he is restricted to stationary situations to operate unnoticed, i.e., the AD vehicle doesn't move and thus the scenery captured by the camera sensor only slightly changes. 
Because of \ifthenelse{\equal{\changesred}{1}}{\red{this operational constraint}}{this operational constraint}, we call this method stationary segmentation mask method (SSMM).
The second method modifies the CNN's original semantic segmentation mask $\boldsymbol{m}\!\left( \boldsymbol{x} \right)$ by replacing a predefined objective class $o = m_i\!\left( \boldsymbol{x} \right)\in\mathcal{S}$ at each corresponding pixel position $i\in\mathcal{I}_o\!\left( \boldsymbol{x} \right)\subset\mathcal{I}$ by the spatial nearest-neighbor class $n_i\!\left( \boldsymbol{x} \right) = m_j\!\left( \boldsymbol{x} \right) \in\mathcal{S}$, with $m_j\!\left( \boldsymbol{x} \right)\ne m_i\!\left( \boldsymbol{x} \right)$.
Here, $\mathcal{I}_o\!\left( \boldsymbol{x} \right)$ is the set of all pixel positions, where $m_i\!\left( \boldsymbol{x} \right)=o$ holds.
By completely removing the objective class $o$ from the semantic segmentation mask, we obtain
\begin{equation}
m_i^{\text{DNNM}}\!\left( \boldsymbol{x} \right) =\begin{cases}
m_i\!\left( \boldsymbol{x} \right), & \text{if $m_i\!\left( \boldsymbol{x} \right)\ne o$}. \\
n_i\!\left( \boldsymbol{x} \right), & \text{otherwise},
\end{cases}
\end{equation}

\ifthenelse{\equal{\mode}{0}}{}{
	\begin{figure}[t!]
		\centering
		\includegraphics[width=\columnwidth]{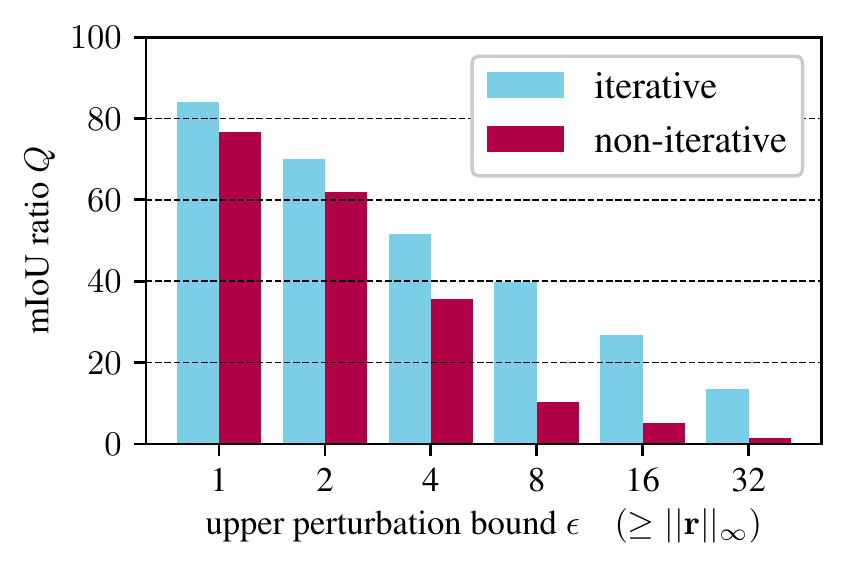}
		\caption{Adversarial attacks on the ICNet using the iterative or the non-iterative least-likely class method (LLCM) from Kurakin et al.~\cite{Kurakin2017a} on the Cityscapes validation set with different values for $\epsilon$, an upper bound of the $l_\infty$-norm of the adversarial perturbation $\boldsymbol{r}$. We set $\lambda=\epsilon$ for the non-iterative LLCM and $\lambda=1$ for the iterative LLCM. A lower mIoU ratio $Q$ means a stronger adversarial attack. Note that the non-iterative LLCM appears to be even more aggressive than the iterative LLCM.}\vspace{\myvspace}
		\label{fig:adv_llm_curves}
\end{figure}}

\ifthenelse{\equal{\mode}{0}}{}{
	\begin{figure*}[t!]
		\includegraphics[width=0.245\textwidth, height=0.125\textwidth]
		{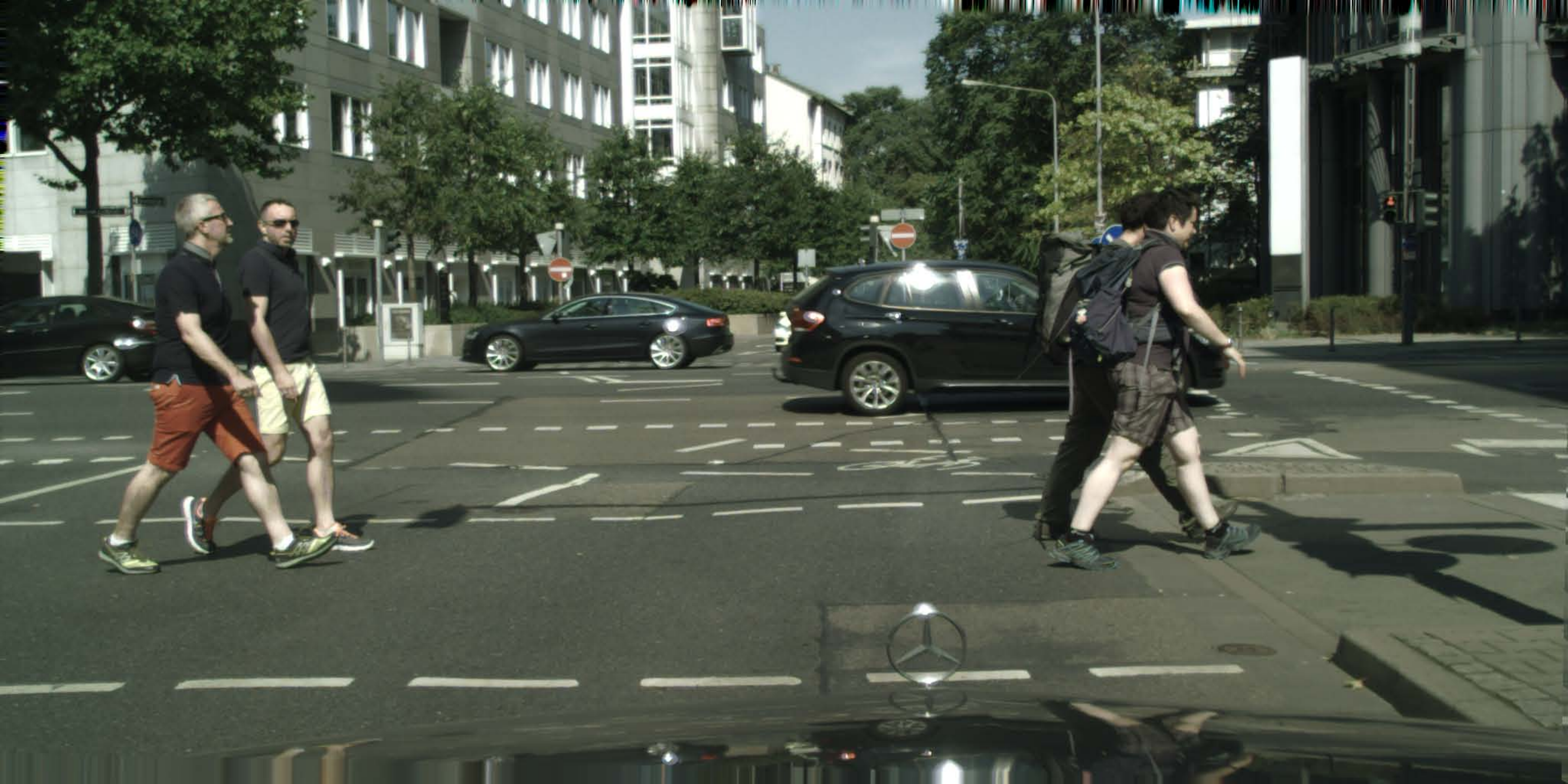}
		\vspace{0.1cm}
		\includegraphics[width=0.245\textwidth, height=0.125\textwidth]
		{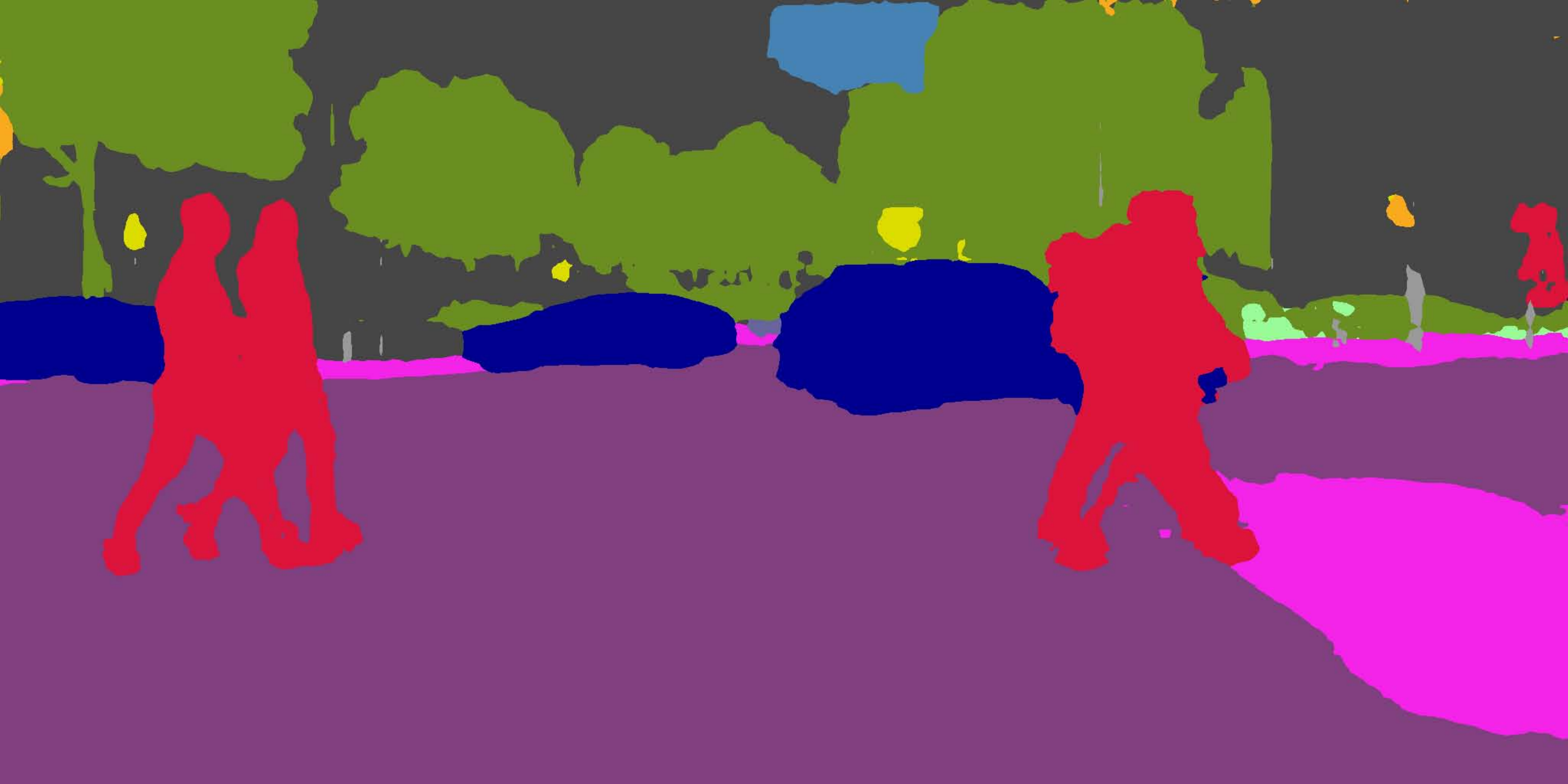}
		\includegraphics[width=0.245\textwidth, height=0.125\textwidth]
		{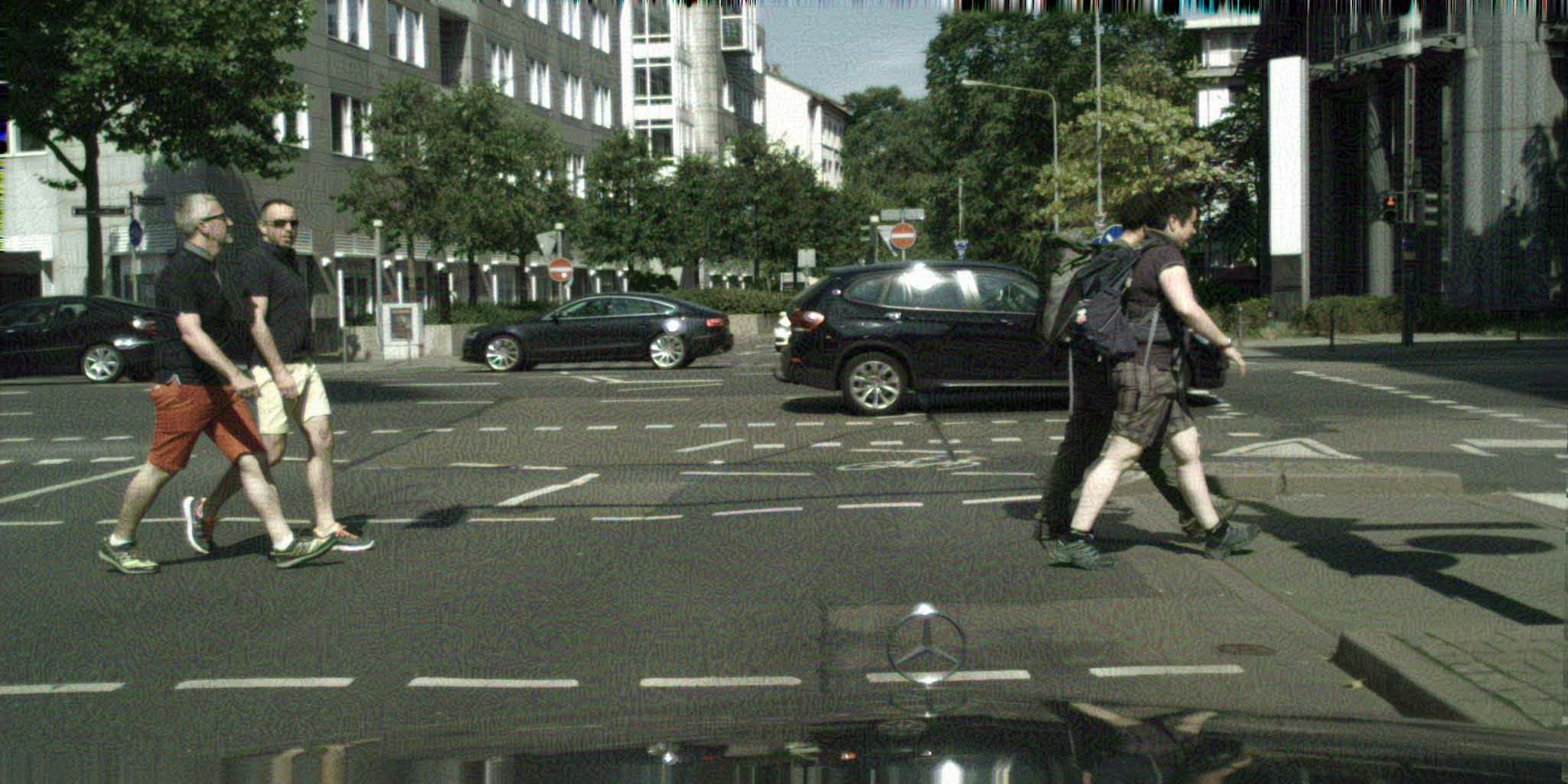}
		\includegraphics[width=0.245\textwidth, height=0.125\textwidth]
		{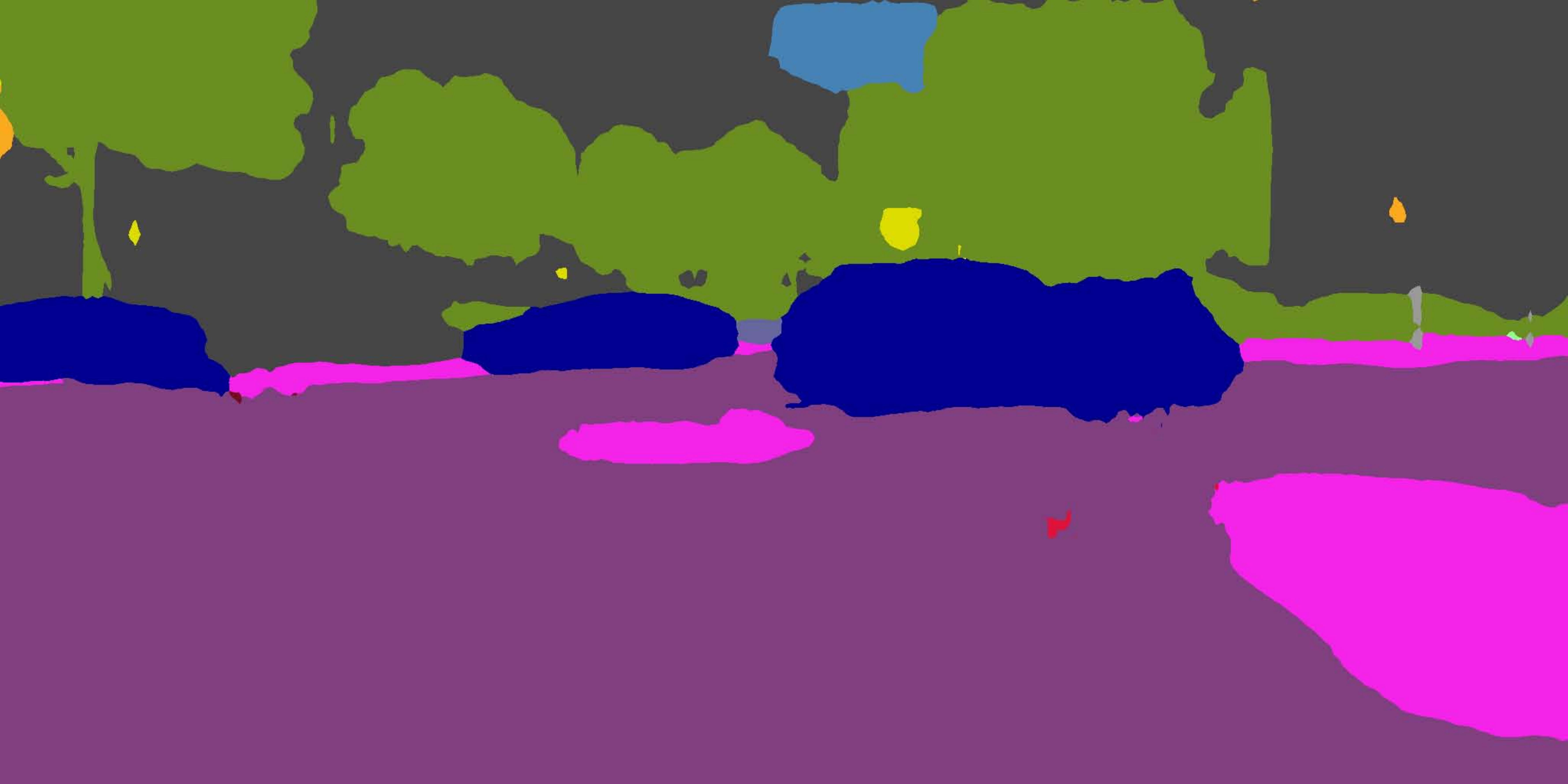}
		
		\includegraphics[width=0.245\textwidth, height=0.125\textwidth]
		{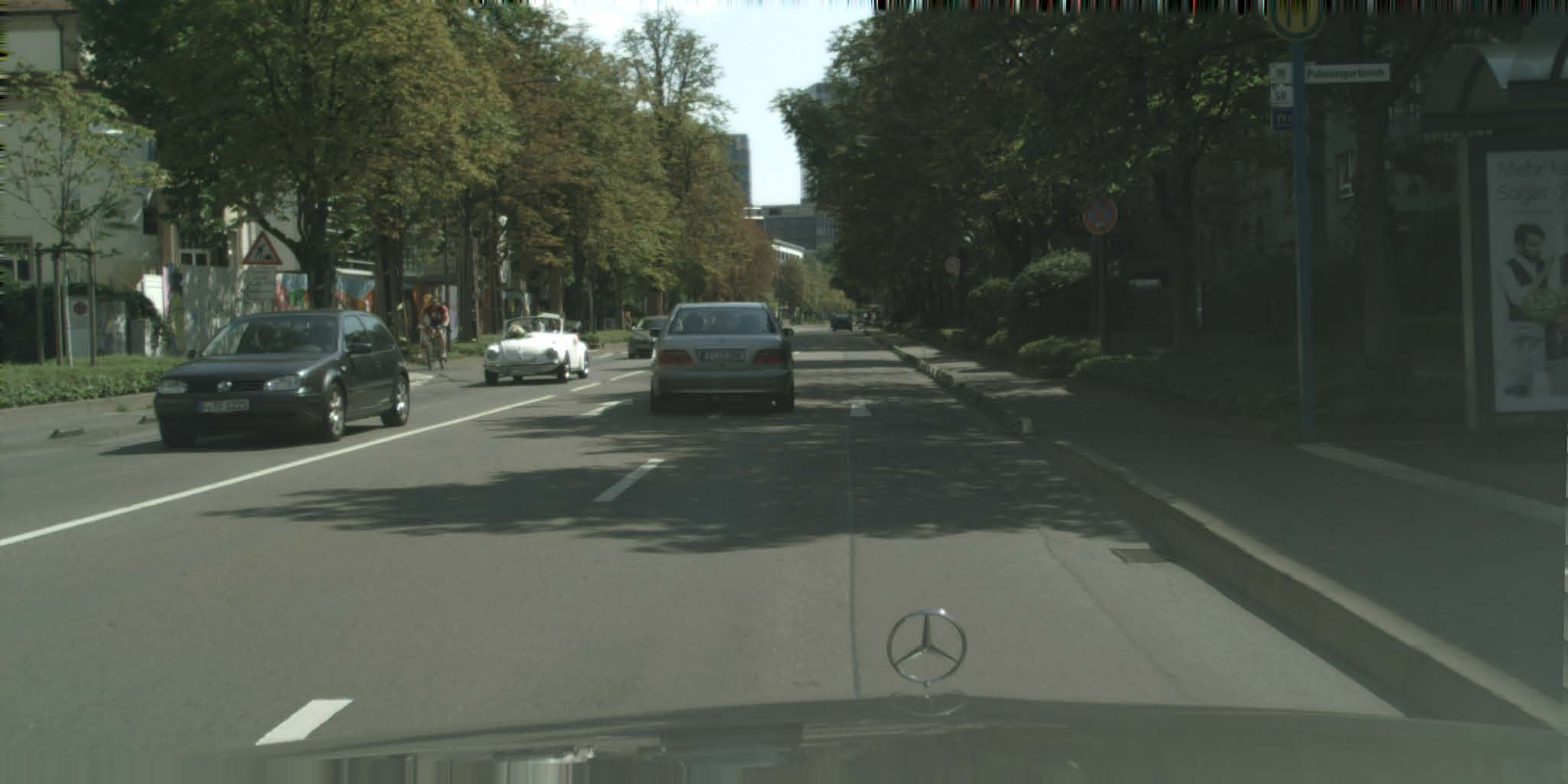}\vspace{0.1cm}
		\includegraphics[width=0.245\textwidth, height=0.125\textwidth]
		{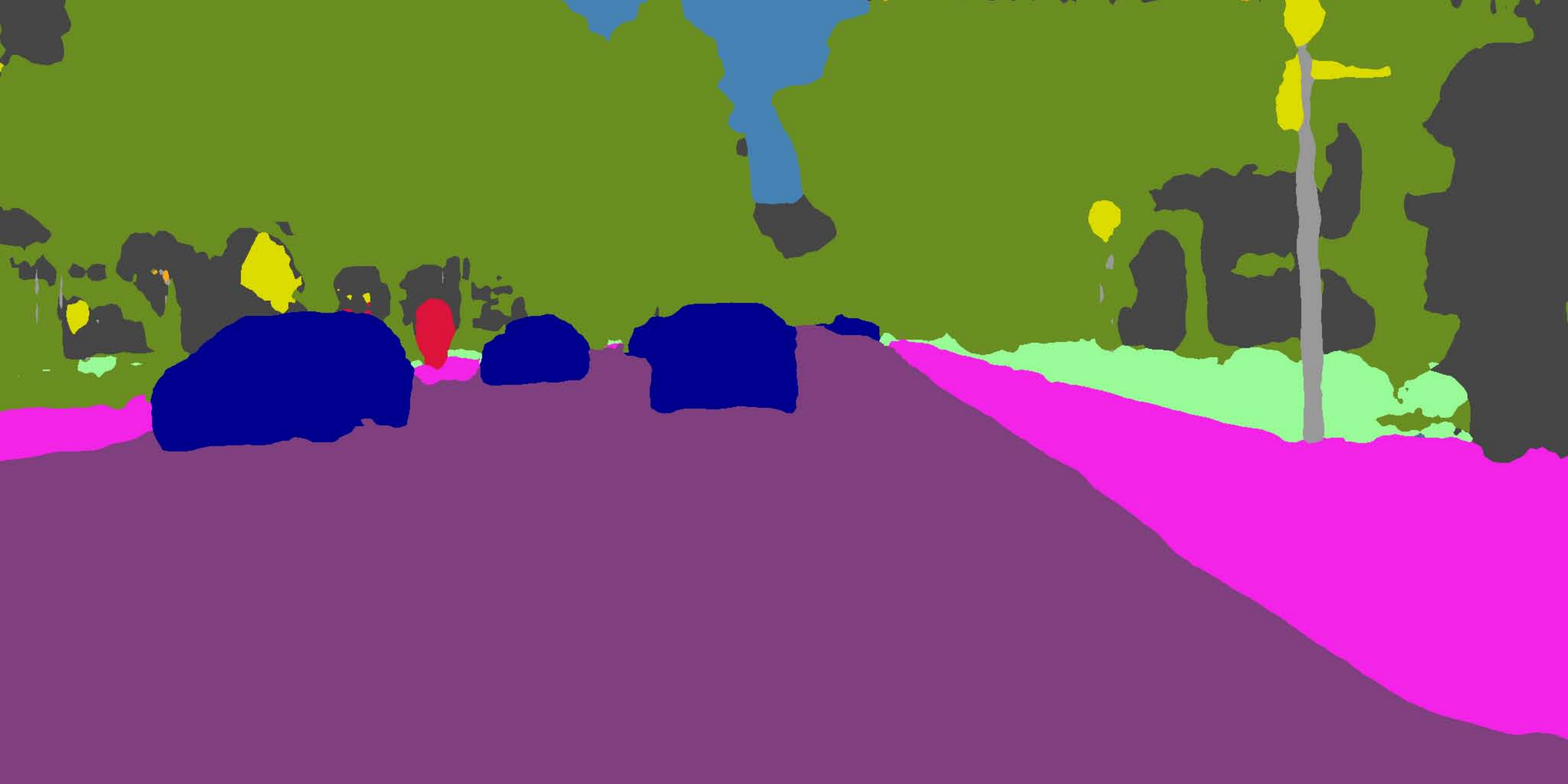}
		\includegraphics[width=0.245\textwidth, height=0.125\textwidth]
		{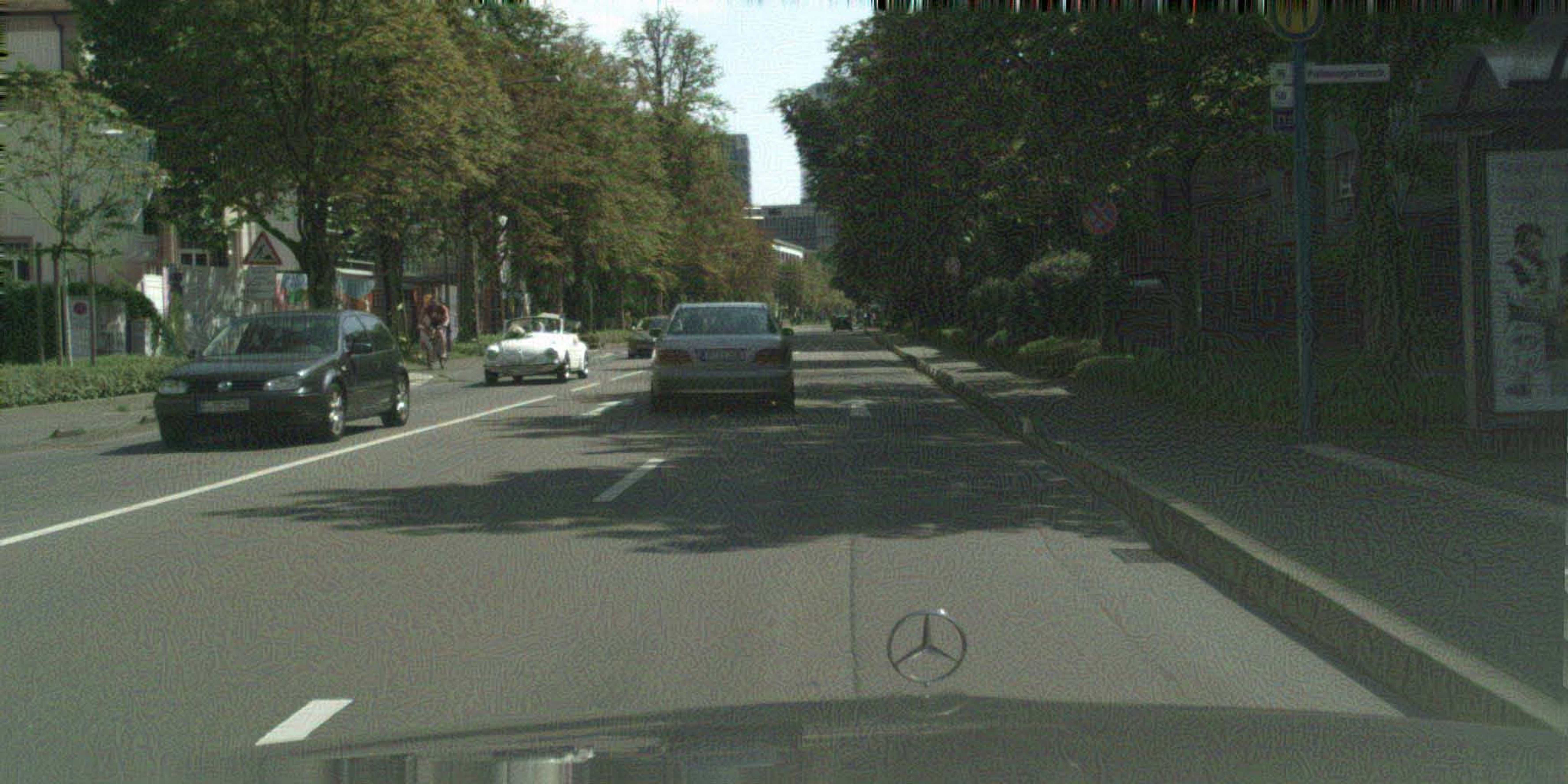}
		\includegraphics[width=0.245\textwidth, height=0.125\textwidth]
		{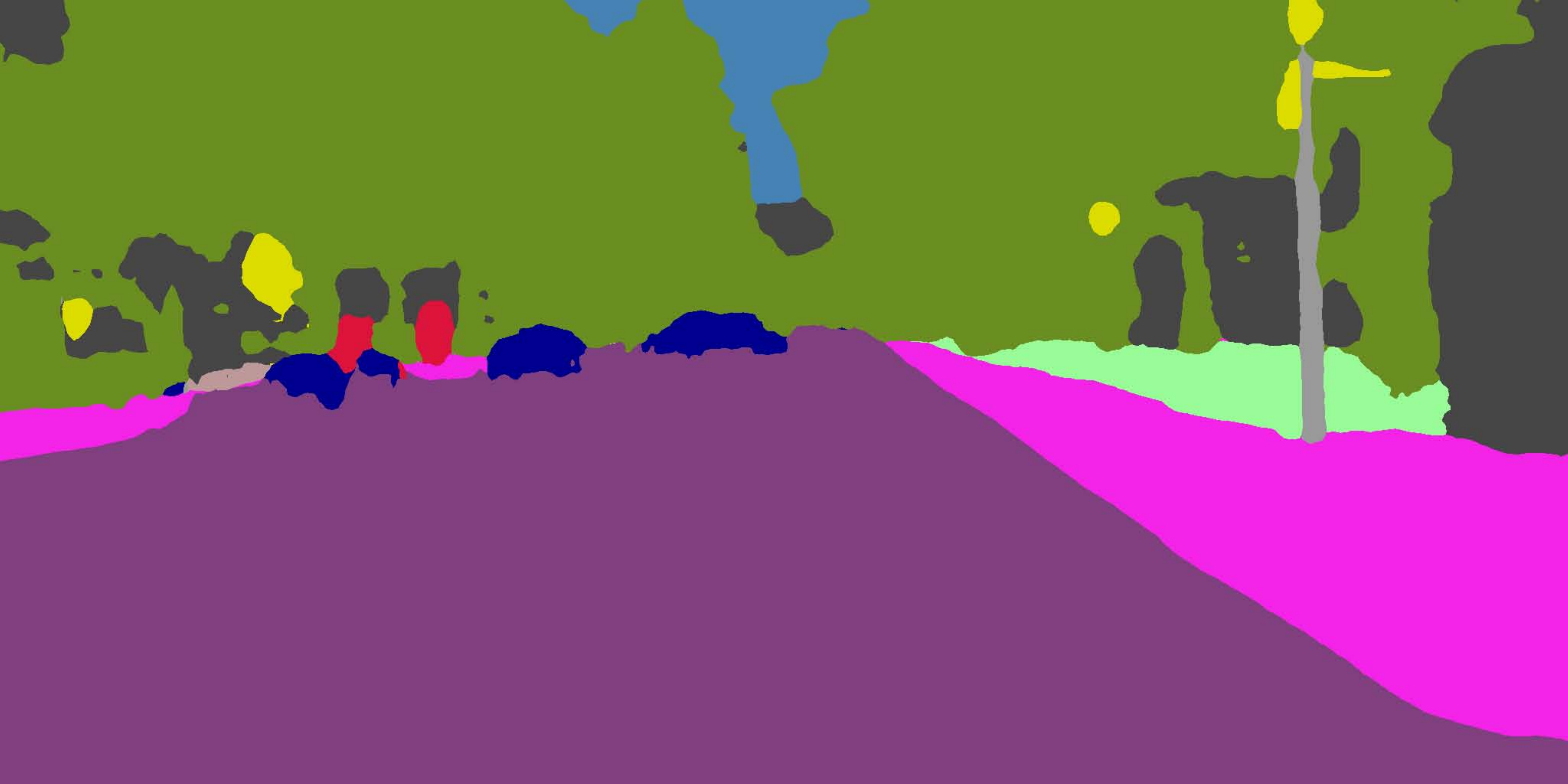}
		
		\begin{tabu} to \textwidth {X[c]X[c]X[c]X[c]}
			{\small (a)} & (b) & (c) & (d)
		\end{tabu}
		
		\caption[toc entry]{Adversarial attacks on the ICNet using the dynamic nearest neighbor method (DNNM) \cite{Metzen2017} on \protect\ifthenelse{\equal{\changesred}{1}}{\red{
					two example images, one with pedestrians and one with cars, from
			}}
			{two example images, one with pedestrians and one with cars, from
			}the Cityscapes validation set. The adversarial examples aim at removing pedestrians (first row) and cars (second row) from the scene. (a) clean input image, (b) semantic segmentation output on clean input image, (c) adversarial example created by DNNM, and (d) semantic segmentation output on adversarial example created by DNNM.}\vspace{\myvspace}
		\label{fig:adv_metzen_attack}
\end{figure*}}

with $m^{\text{DNNM}}_i\!\left( \boldsymbol{x} \right)$ being the new target class at pixel position $i\in\mathcal{I}$.
Metzen et al.~suggested to use the Euclidean distance of two pixel positions $i$ and $j$ in order to find the nearest-neighbor class \ifthenelse{\equal{\changesred}{1}}{\red{satisfying}}{satisfying} $n_i\!\left( \boldsymbol{x}\right) = m_j\!\left( \boldsymbol{x} \right)\ne m_i\!\left( \boldsymbol{x} \right)=o$.
In contrast to SSMM, the created realistically looking fake semantic segmentation mask using this method is now unique for each real semantic segmentation output.
Additionally, specific properties, such as the correlation between two consecutive real semantic segmentation outputs, are transferred to the created fake ones. 
Altogether, a possible attacker is able to create a sequence of correlated realistically looking fake semantic segmentation masks making this kind of attack suitable for situations, where the respective AD vehicle moves.
Due to these properties we call this method dynamic nearest neighbor method (DNNM).
The application of DNNM has the potential to create safety-relevant perception errors for AD.
This can be seen in Fig.~\ref{fig:adv_metzen_attack}, where DNNM is used to remove pedestrians or cars from the scene.
The adversarial examples were created by setting $\epsilon=10$ followed by the same procedure as with iterative LLCM.
Astonishingly, the semantic classes different from the objective class are completely preserved and the nearest neighbor class seems to be a good estimate for the regions occluded by the objective class, thereby dangerously providing a plausible but wrong semantic segmentation mask.

\subsection{Universal Adversarial Perturbations}
So far, we \ifthenelse{\equal{\changesred}{1}}{\red{discussed}}{discussed} approaches that generate adversarial perturbations for single input images.
In reality, however, it is hard for a possible attacker to generate adversarial examples for each incoming image of an environment perception system, considering a camera running at 20 fps.
Therefore, in AD applications a special interest lies in single adversarial perturbations being capable of fooling a CNN on a set of input images, e.g., a video sequence.
This class of adversarial perturbations is called universal adversarial perturbation (UAP).

One of the first works towards finding UAPs was done by Moosavi-Dezfooli et al.~\cite{Moosavi-Dezfooli2017}.
Their idea was to find a UAP $\boldsymbol{r}_\text{uni}$ that fools almost all images in some image set $\mathcal{T}$ in an image classification task (again, only one class per image).
To achieve this, they used the DeepFool algorithm in an iterative fashion to solve the optimization problem
\newcommand\mystrut{\rule{0pt}{7.5pt}}
\begin{equation}\label{eq:moos_uni}
s^*\!\left( \boldsymbol{x} + \boldsymbol{r}_\text{uni} \right) \ne s^*\!\left( \boldsymbol{x} \right) \quad \forall \boldsymbol{x}\in \mathcal{T}^{\prime} \subset\mathcal{T},
\end{equation}
with the subset of respective images $\mathcal{T}\mystrut^{\prime}$ for which the CNN is fooled, and the set of all respective images $\mathcal{T}$ the UAP is optimized on.
The UAP is again constrained by
\begin{equation}\label{eq:moos_uni_2}
||\boldsymbol{r}_\text{uni}||_p \le \epsilon,
\end{equation}
with $|| \cdot ||_p$ being the $l_p$-norm of $\boldsymbol{r}_\text{uni}$, and $\epsilon$ being its upper bound.
In their experiments, Moosavi-Dezfooli et al.~obtained the best results setting $p=\infty$ and $\epsilon=10$.
Different from all the attacks shown before, the UAP optimized on $\mathcal{T}$ generalizes well, meaning the UAP can even fool a respective system on a disjoint set of images $\mathcal{V}$, with $\mathcal{V}\cap\mathcal{T}=\emptyset$, on which the UAP was not optimized on.

\ifthenelse{\equal{\mode}{0}}{}
{\begin{figure*}[t!]
		\includegraphics[width=0.245\textwidth, height=0.125\textwidth]{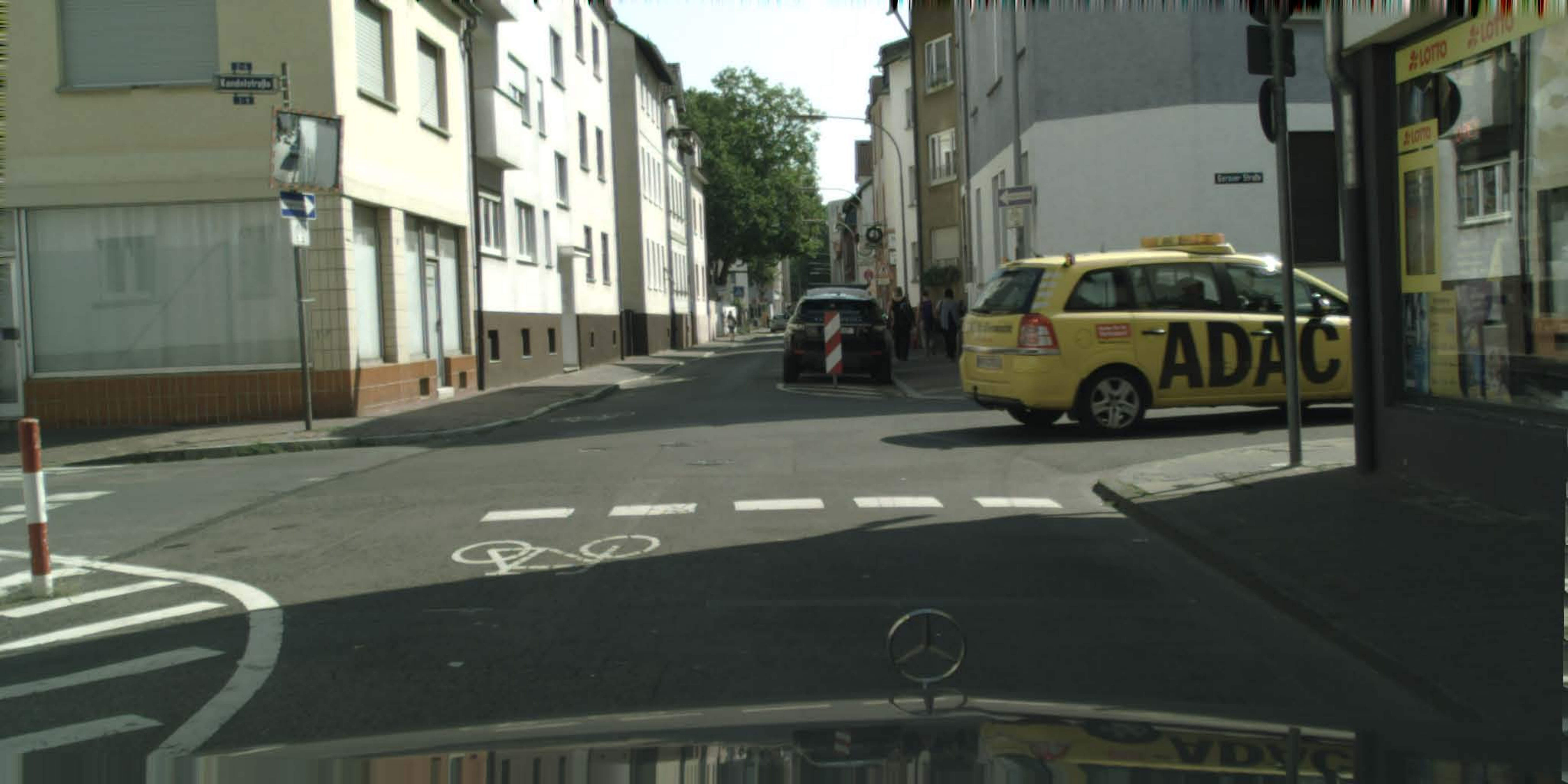}
		\vspace{0.1cm}
		\includegraphics[width=0.245\textwidth, height=0.125\textwidth]{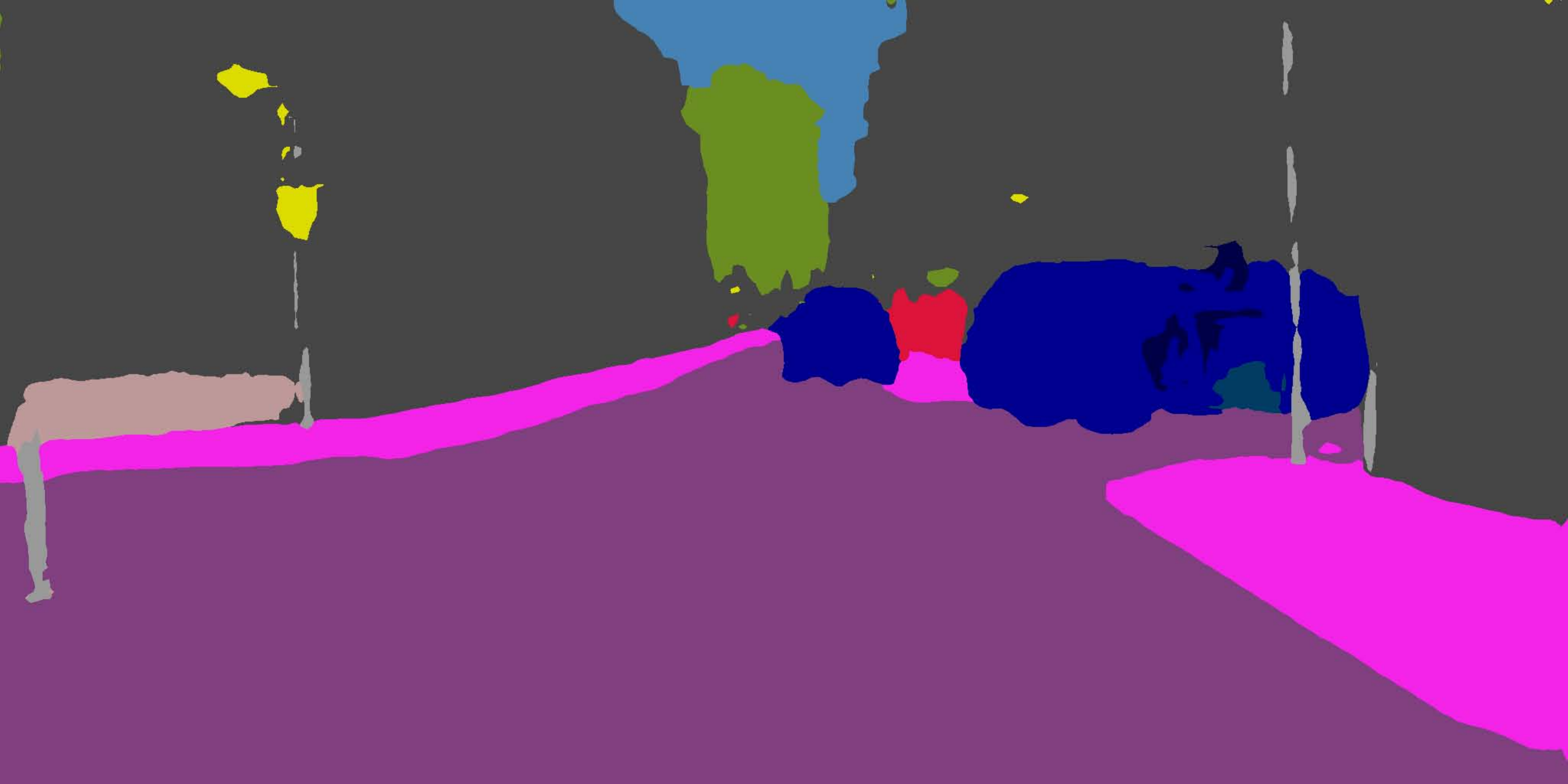}
		\includegraphics[width=0.245\textwidth, height=0.125\textwidth]{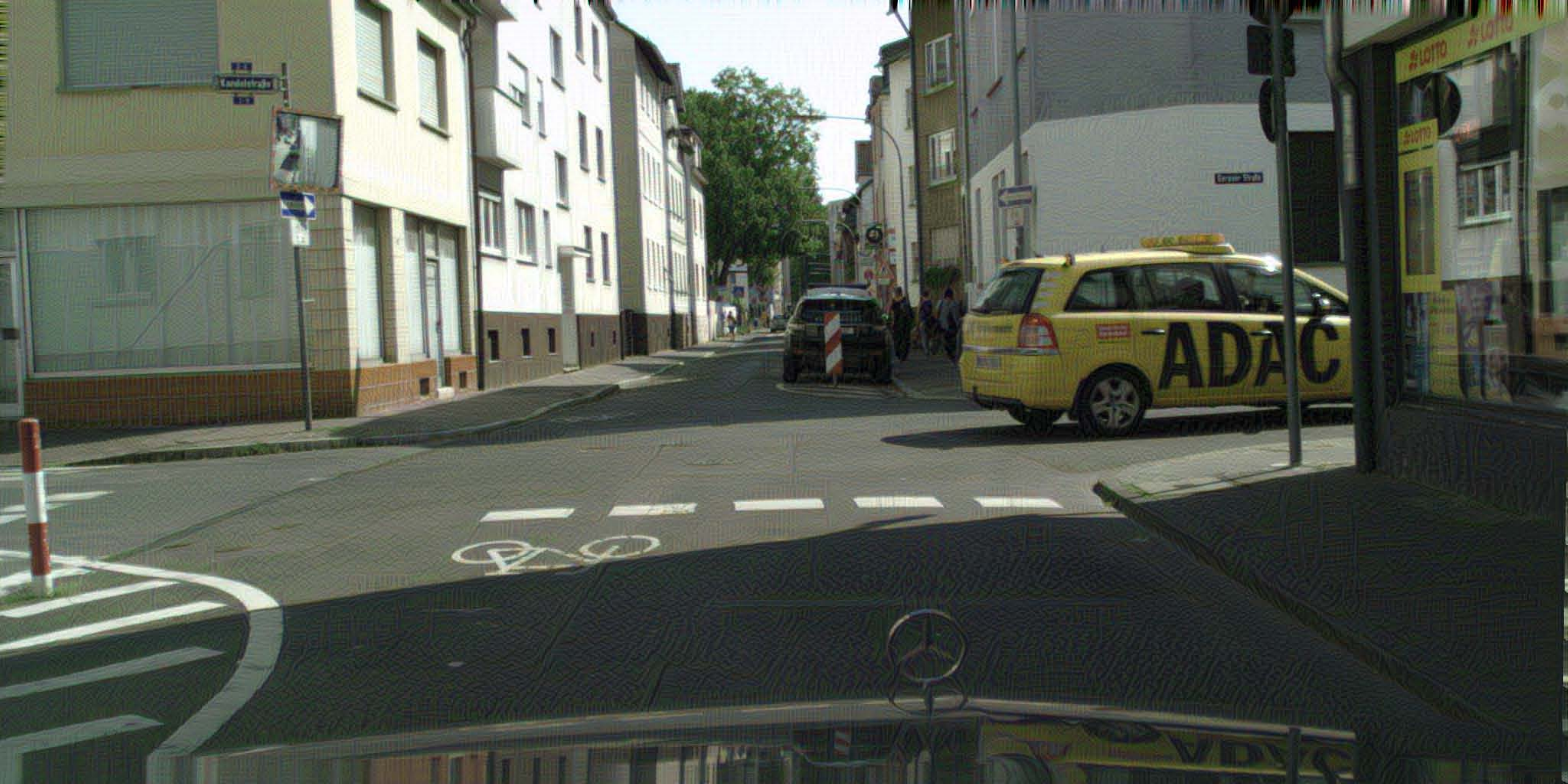}
		\includegraphics[width=0.245\textwidth, height=0.125\textwidth]{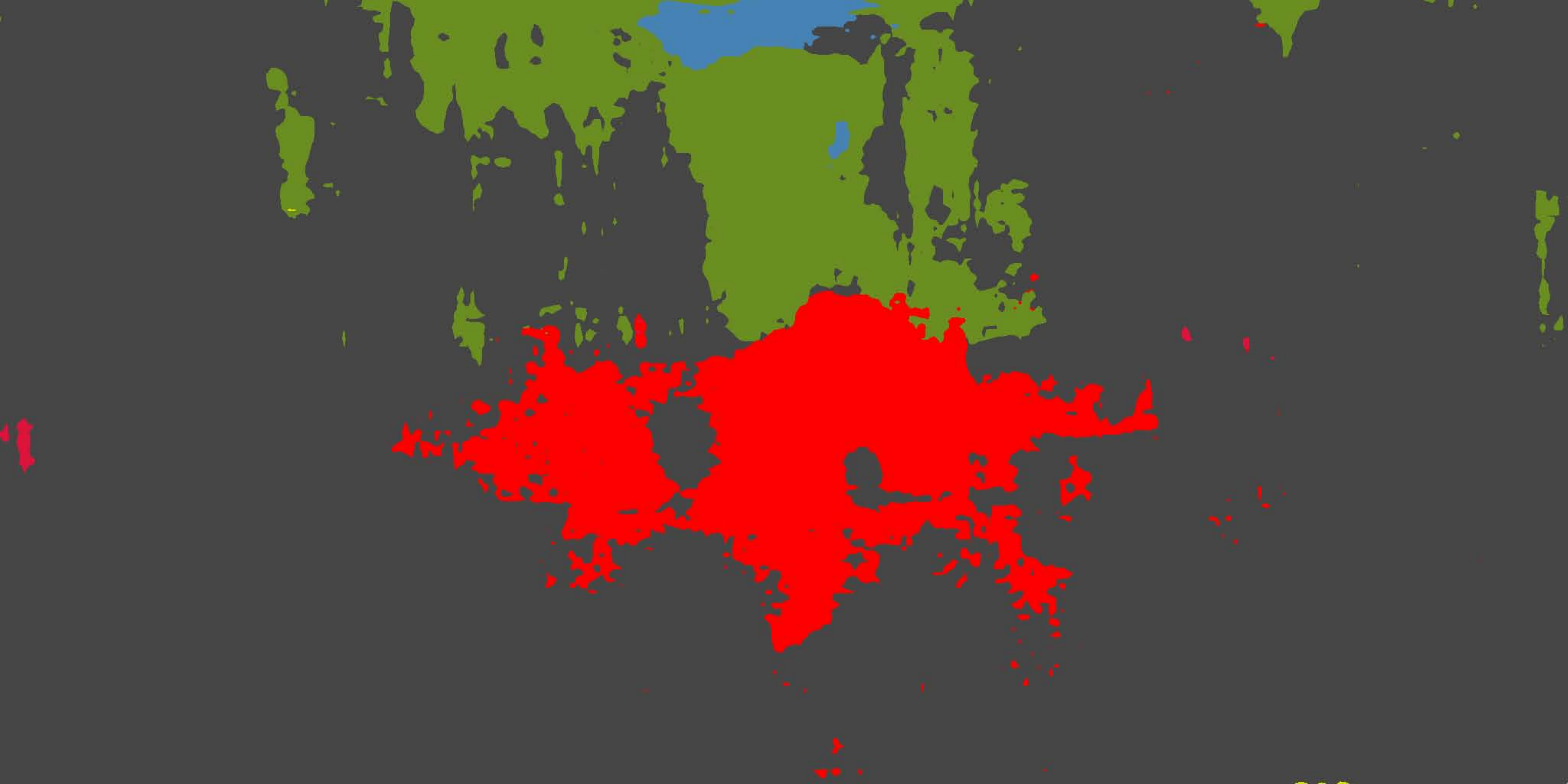}
		
		\includegraphics[width=0.245\textwidth, height=0.125\textwidth]{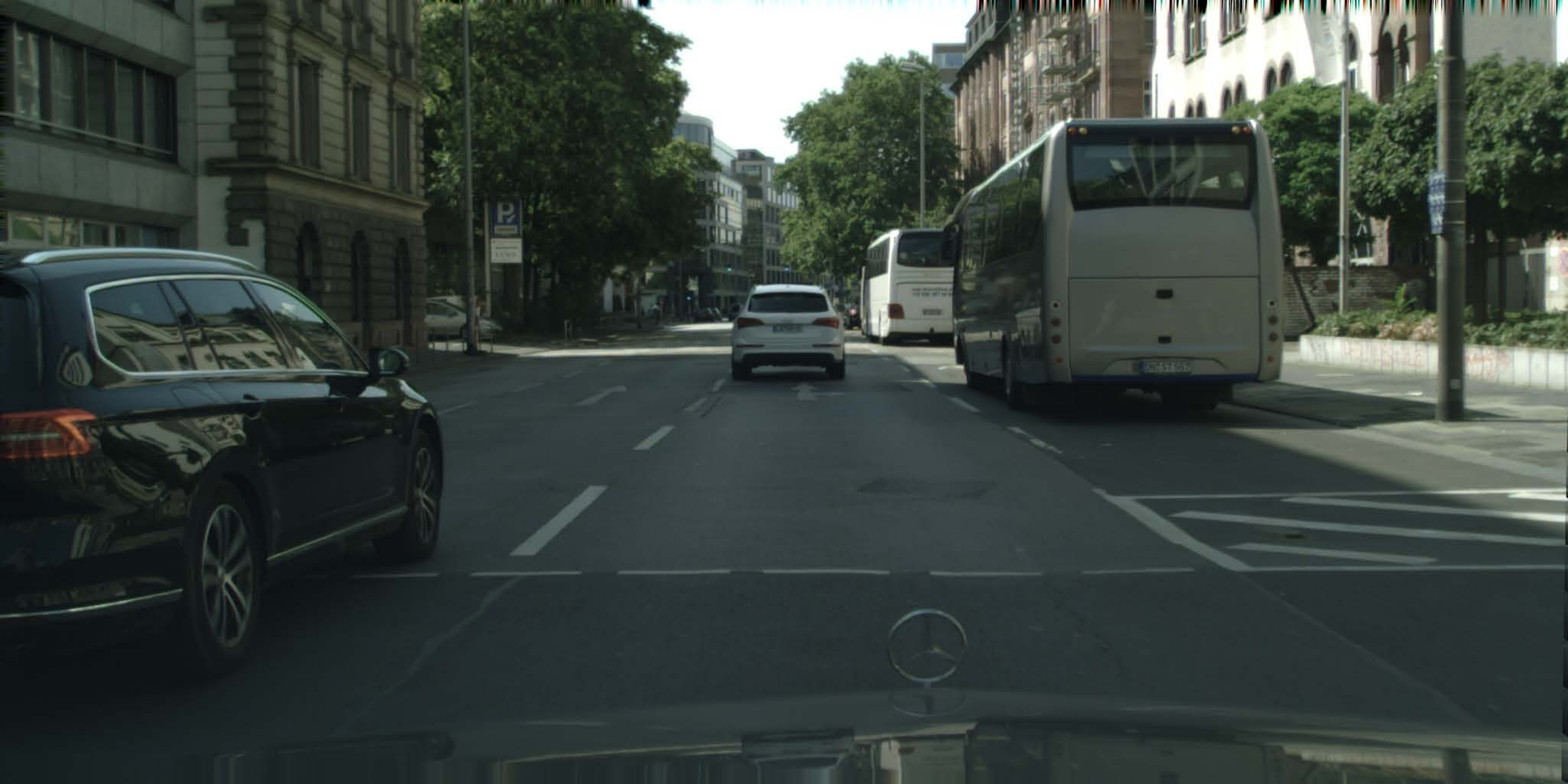}
		\vspace{0.1cm}
		\includegraphics[width=0.245\textwidth, height=0.125\textwidth]{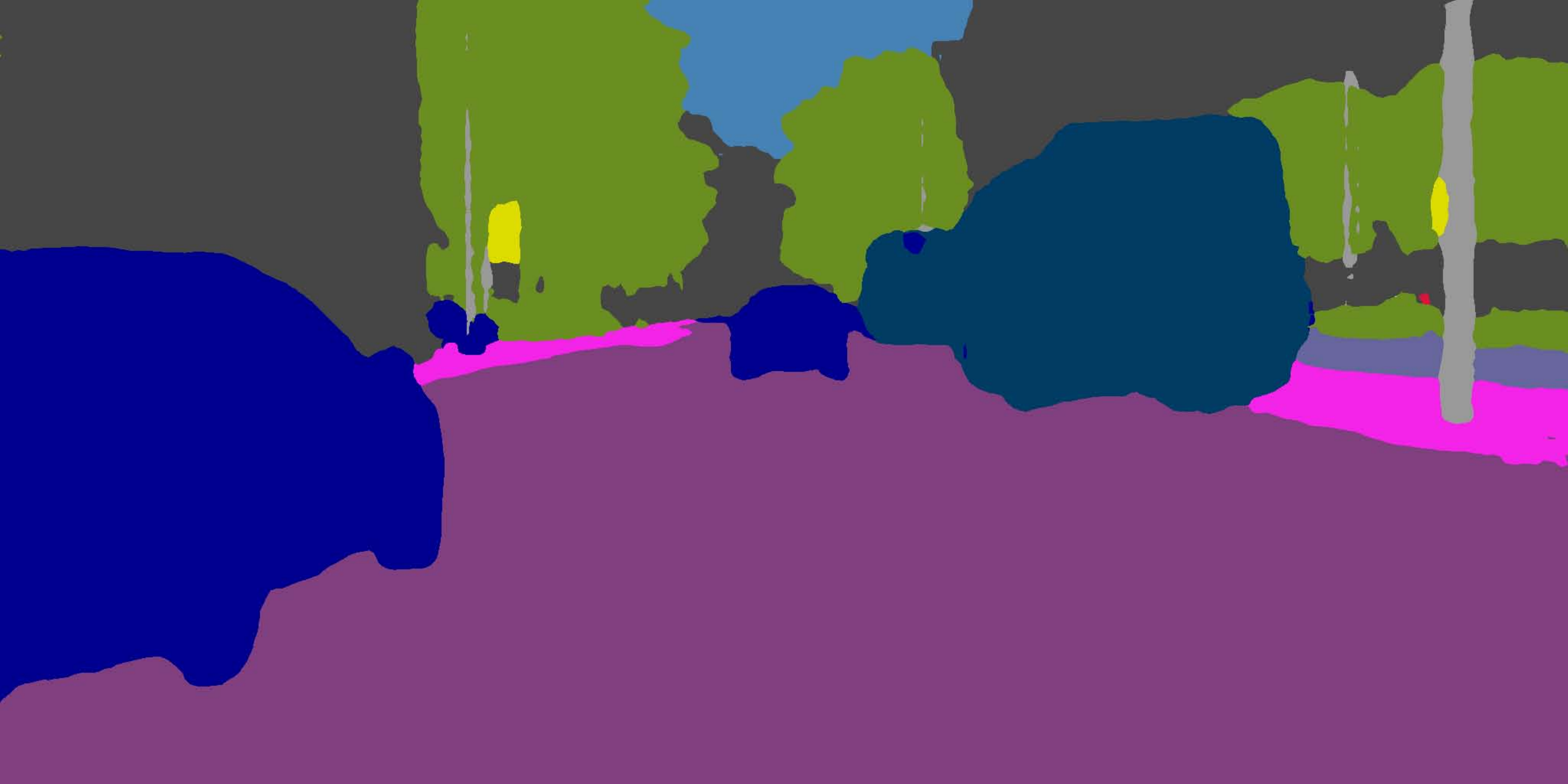}
		\includegraphics[width=0.245\textwidth, height=0.125\textwidth]{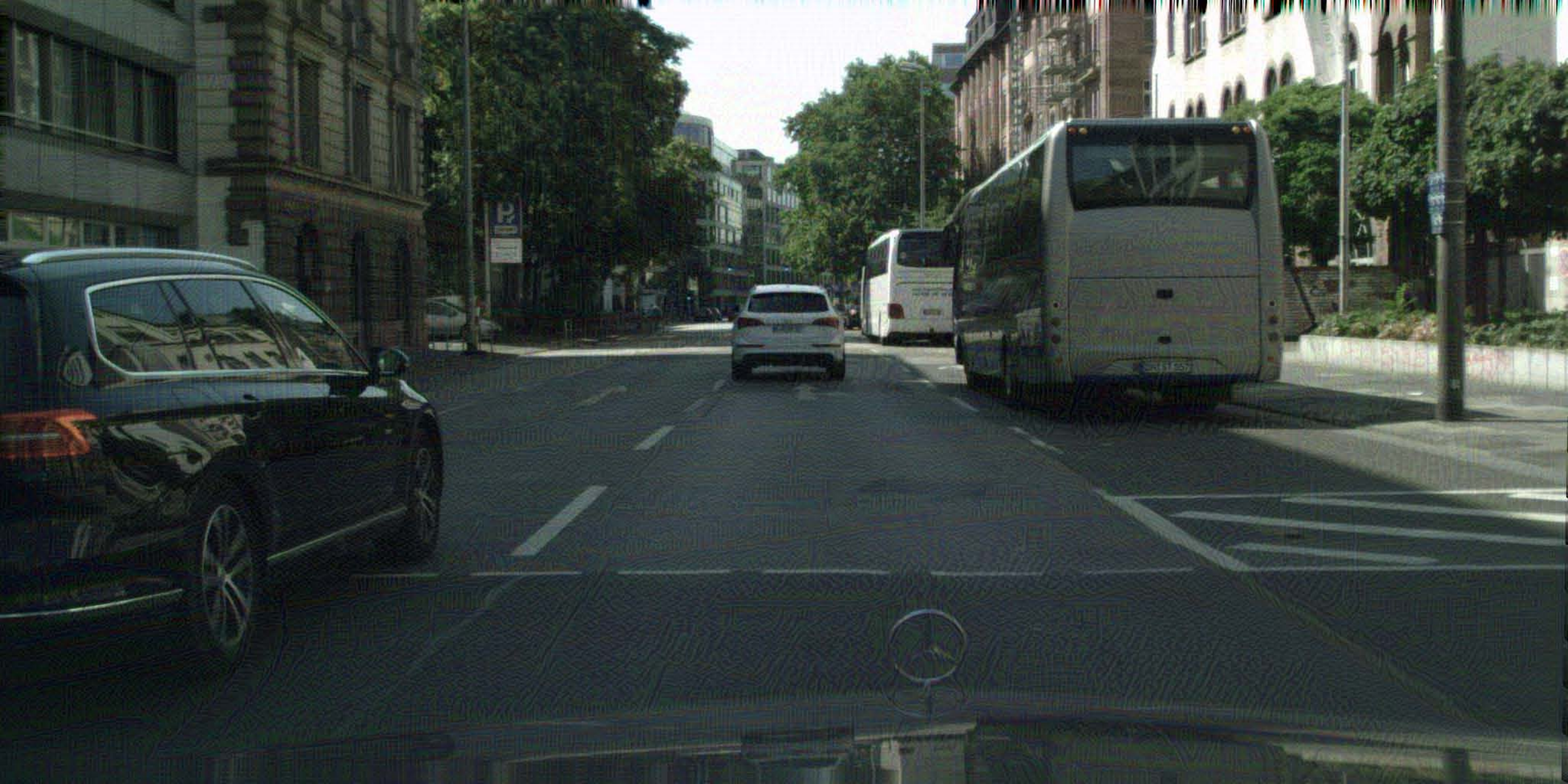}
		\includegraphics[width=0.245\textwidth, height=0.125\textwidth]{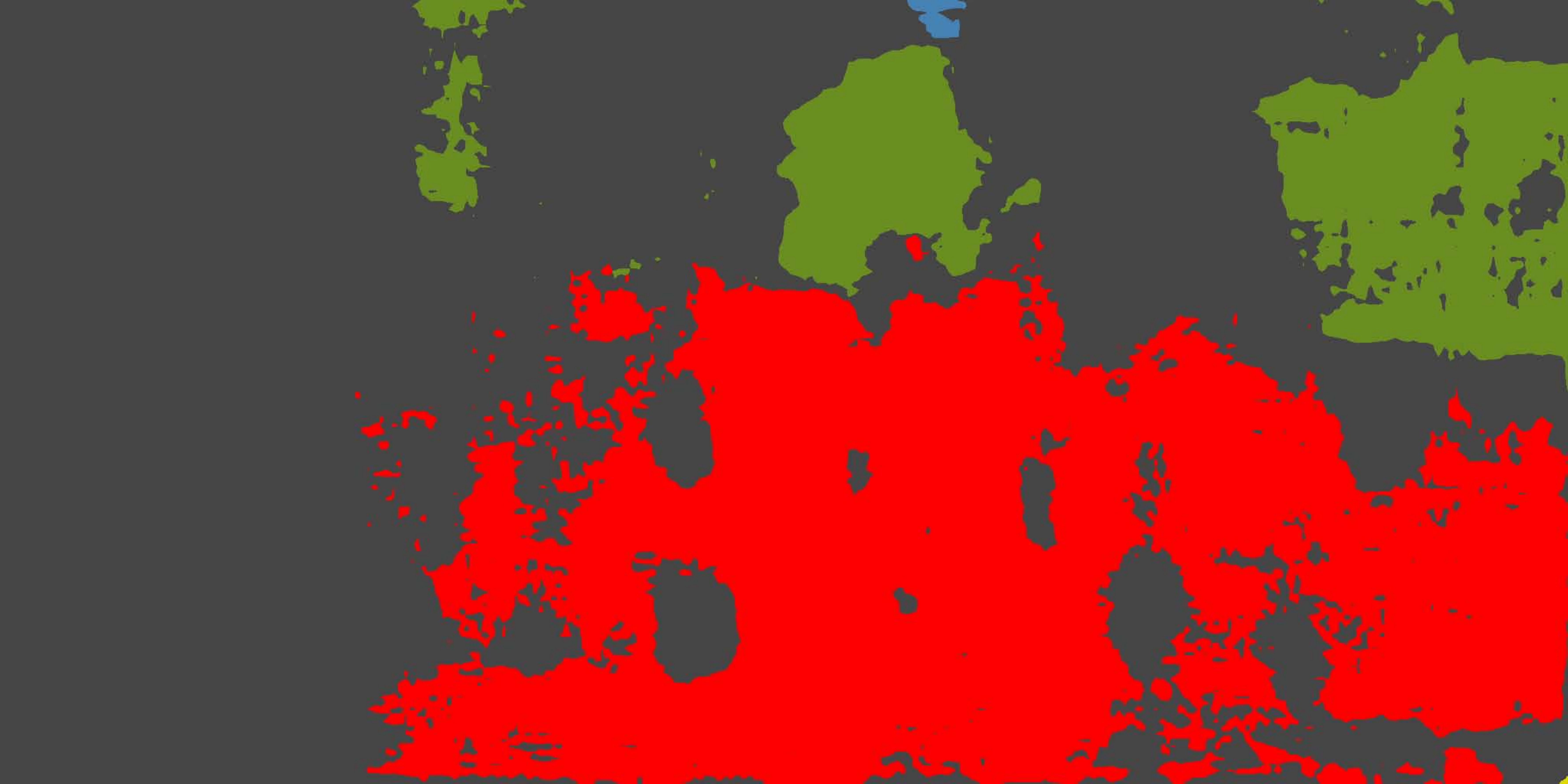}
		
		\includegraphics[width=0.245\textwidth, height=0.125\textwidth]{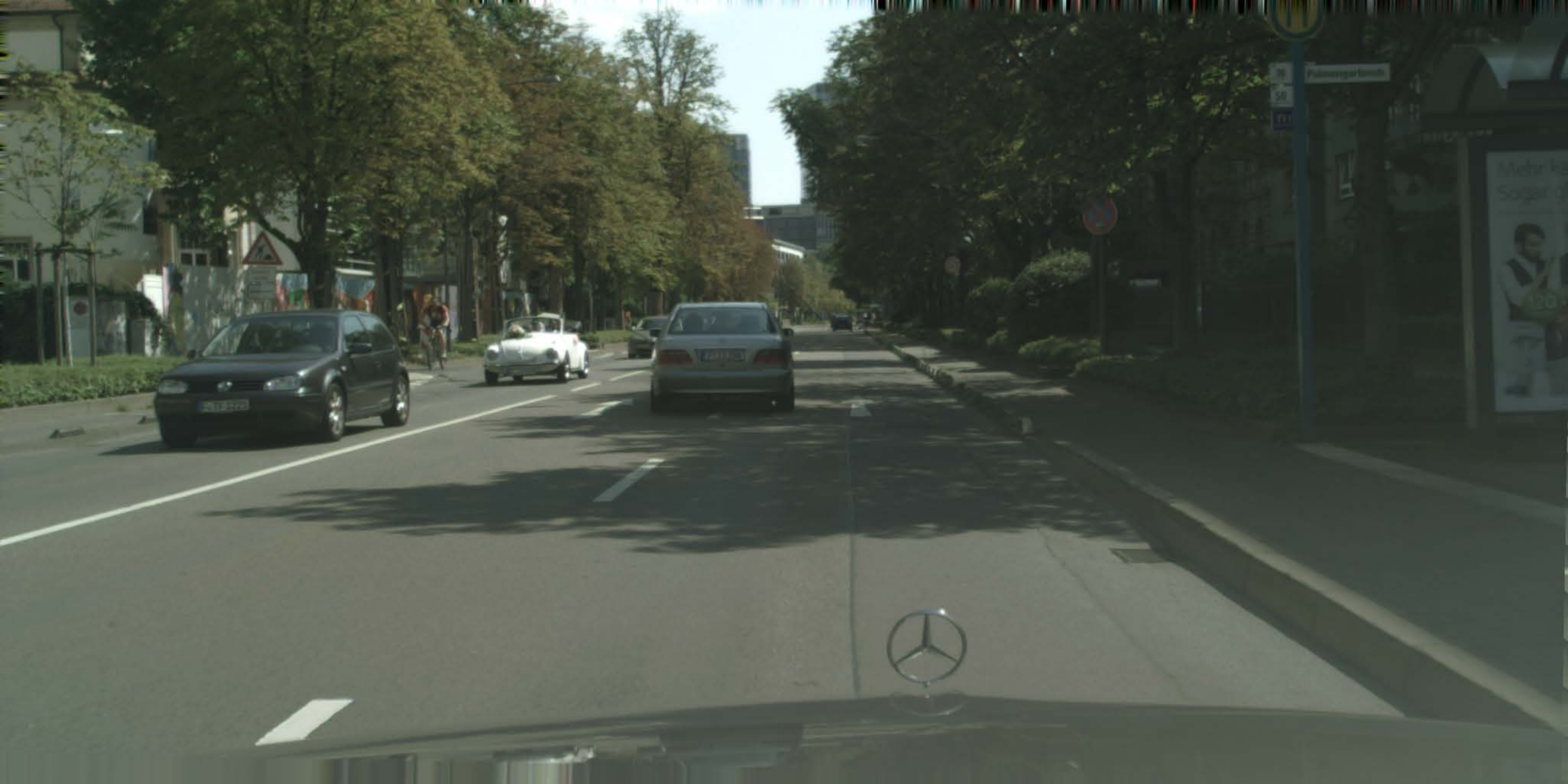}
		\vspace{0.1cm}
		\includegraphics[width=0.245\textwidth, height=0.125\textwidth]{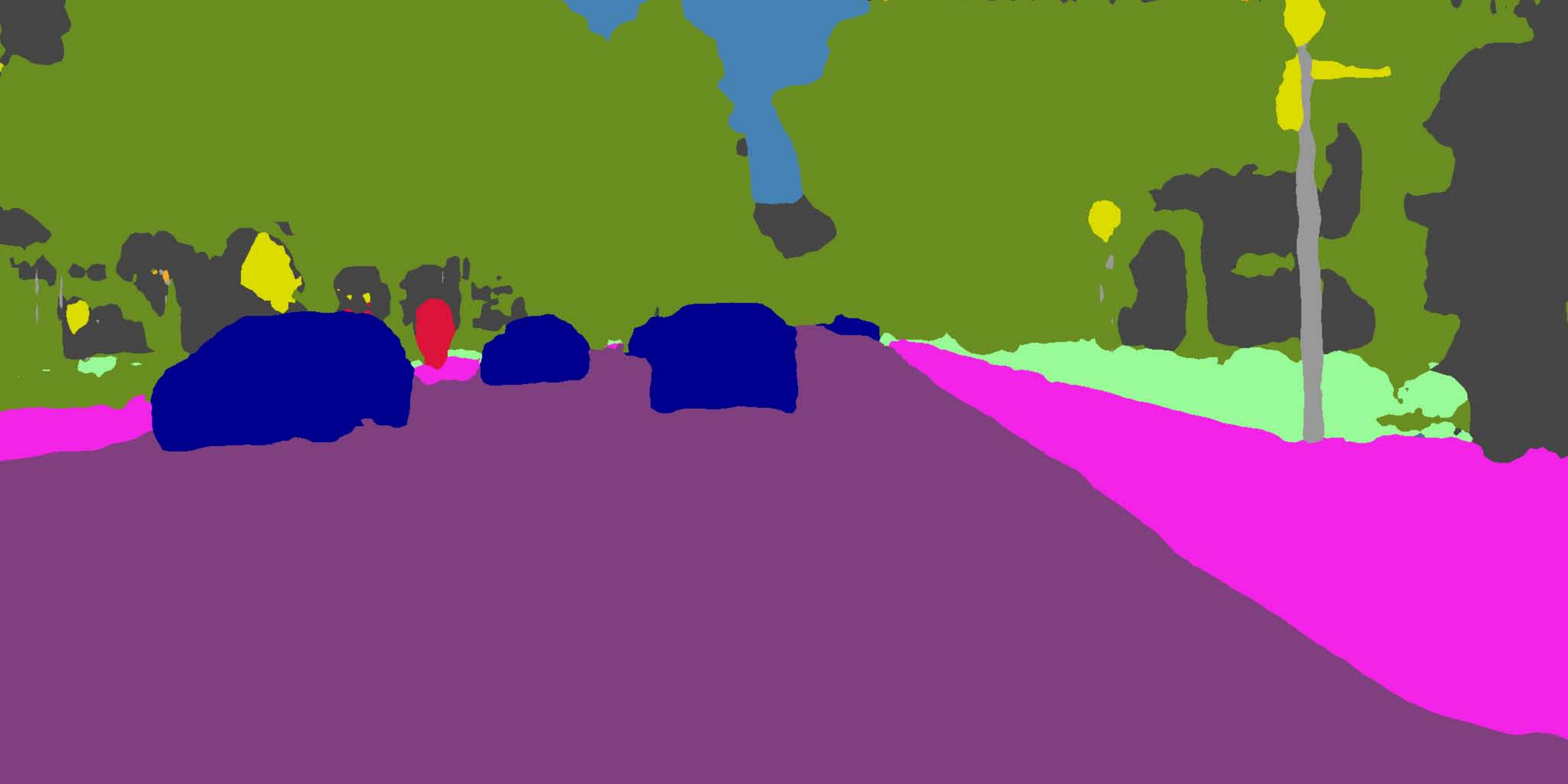}
		\includegraphics[width=0.245\textwidth, height=0.125\textwidth]{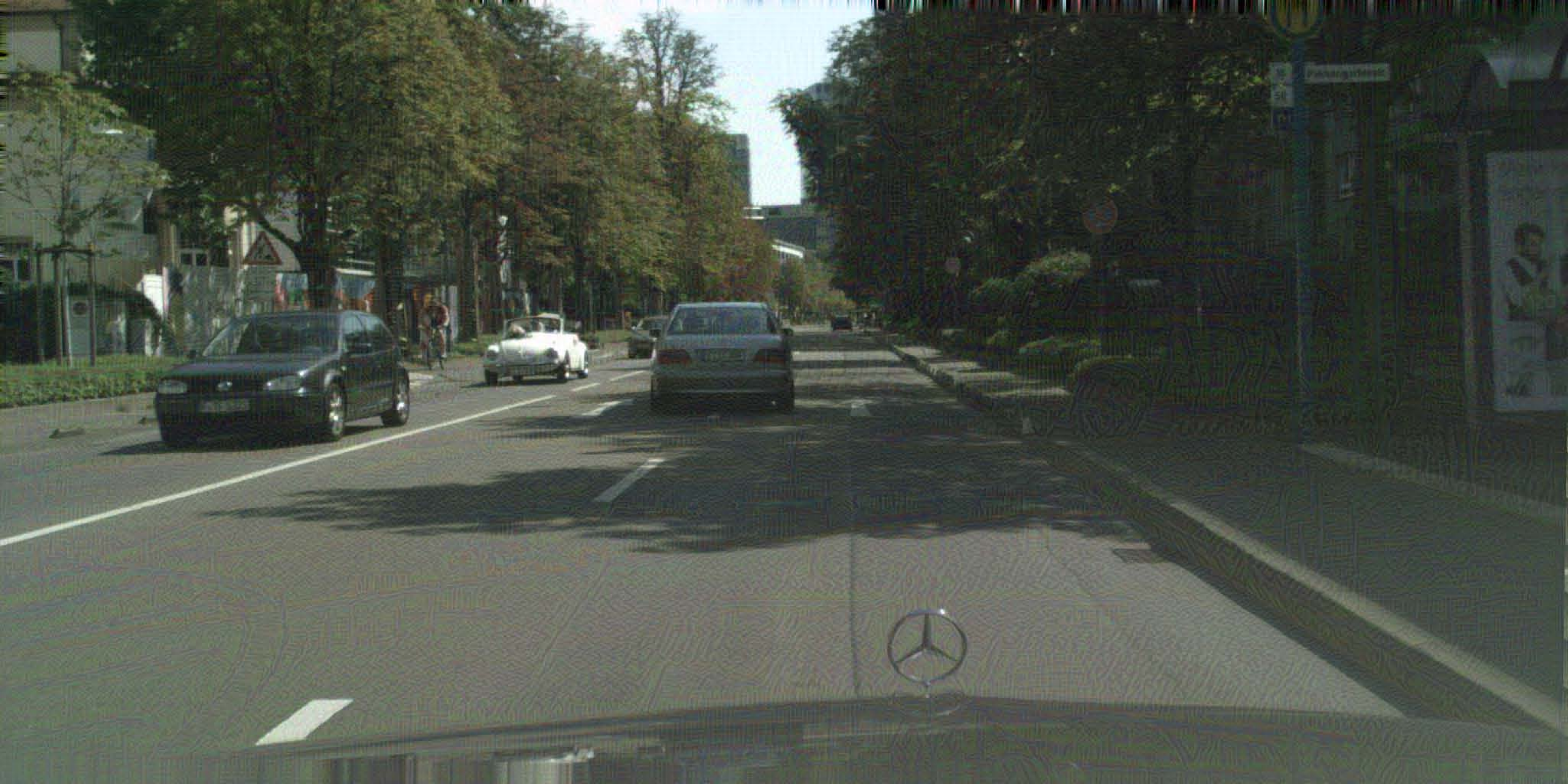}
		\includegraphics[width=0.245\textwidth, height=0.125\textwidth]{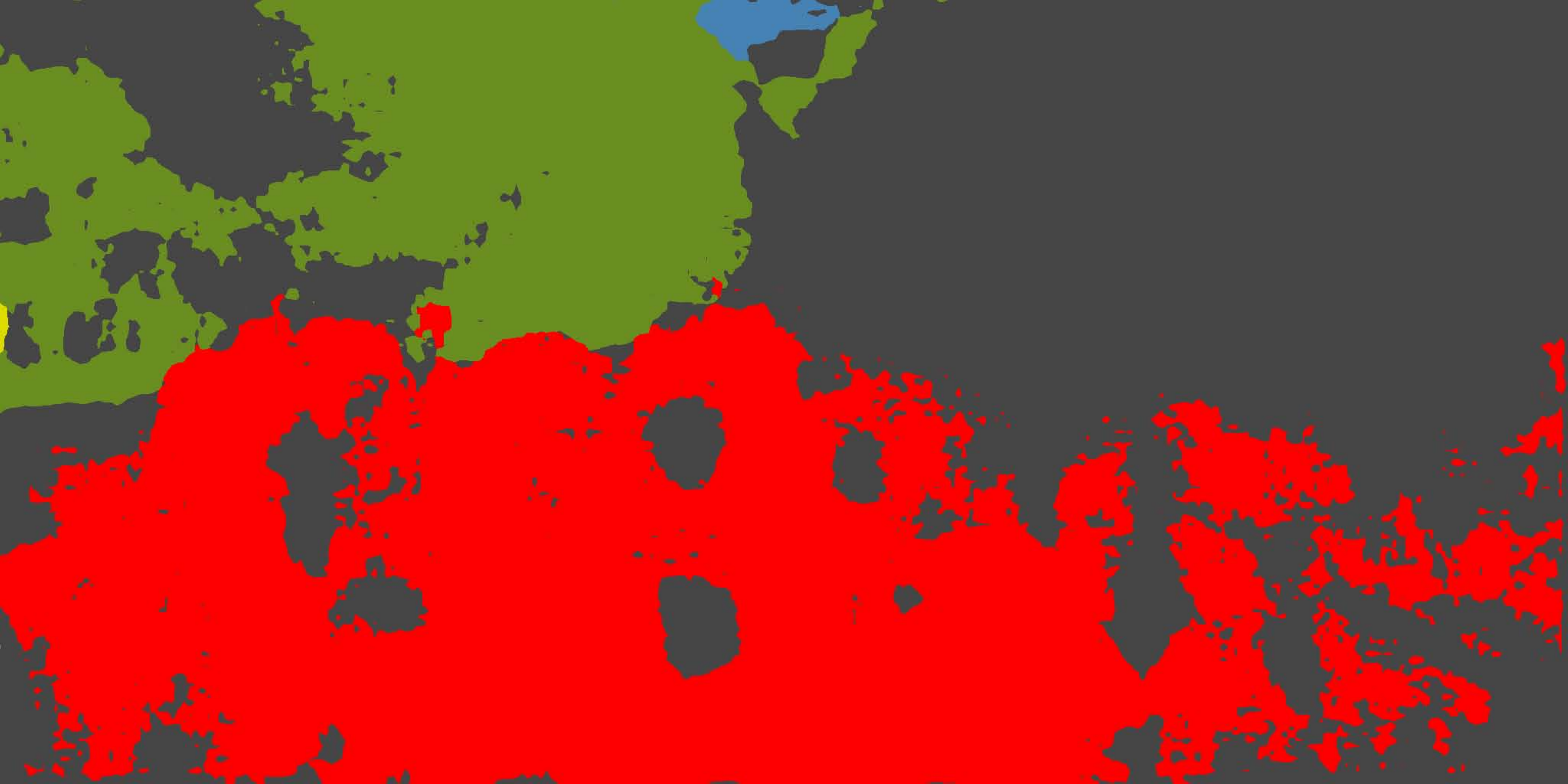}
		
		\includegraphics[width=0.245\textwidth, height=0.125\textwidth]{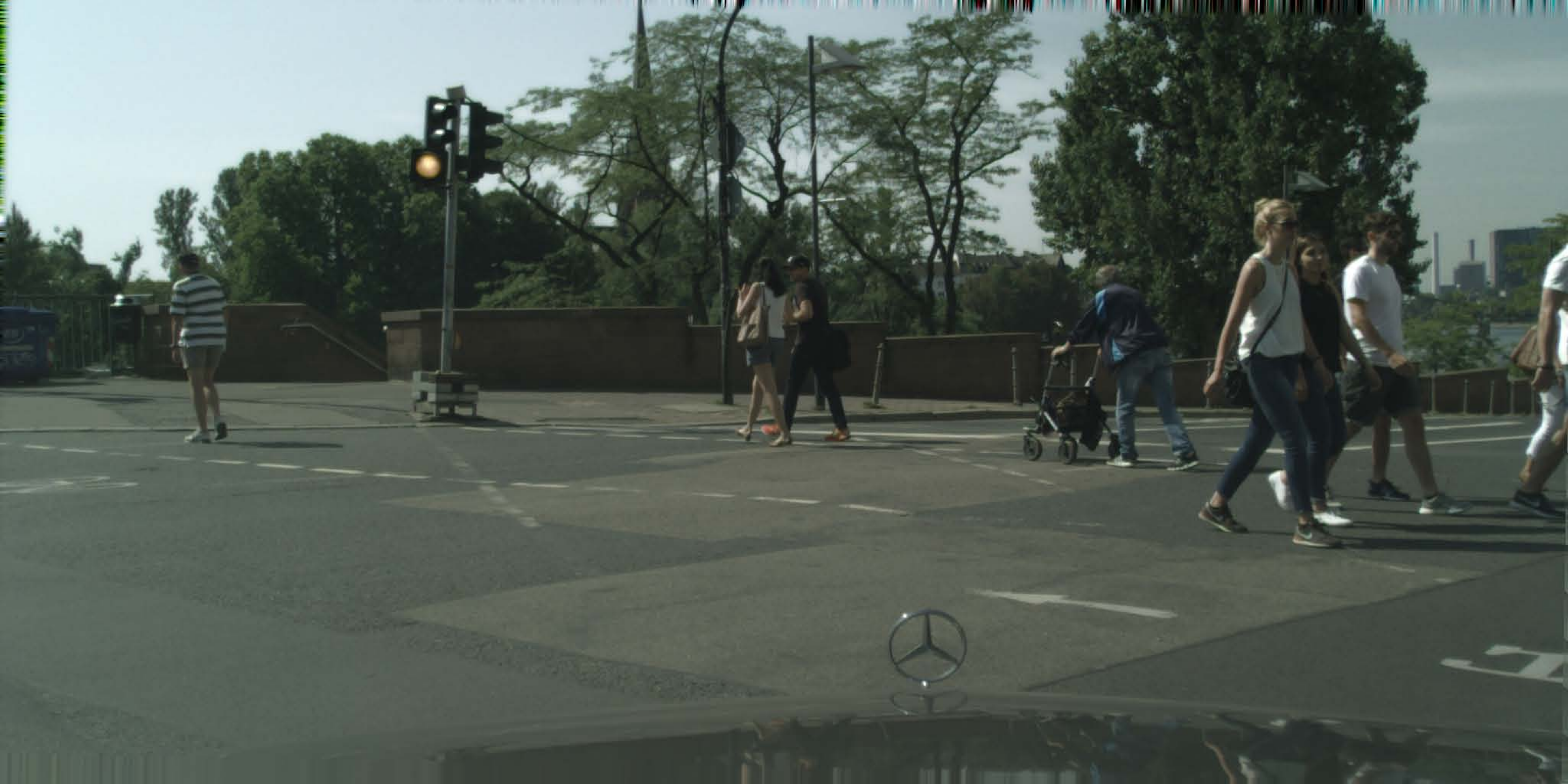}
		\vspace{0.1cm}
		\includegraphics[width=0.245\textwidth, height=0.125\textwidth]{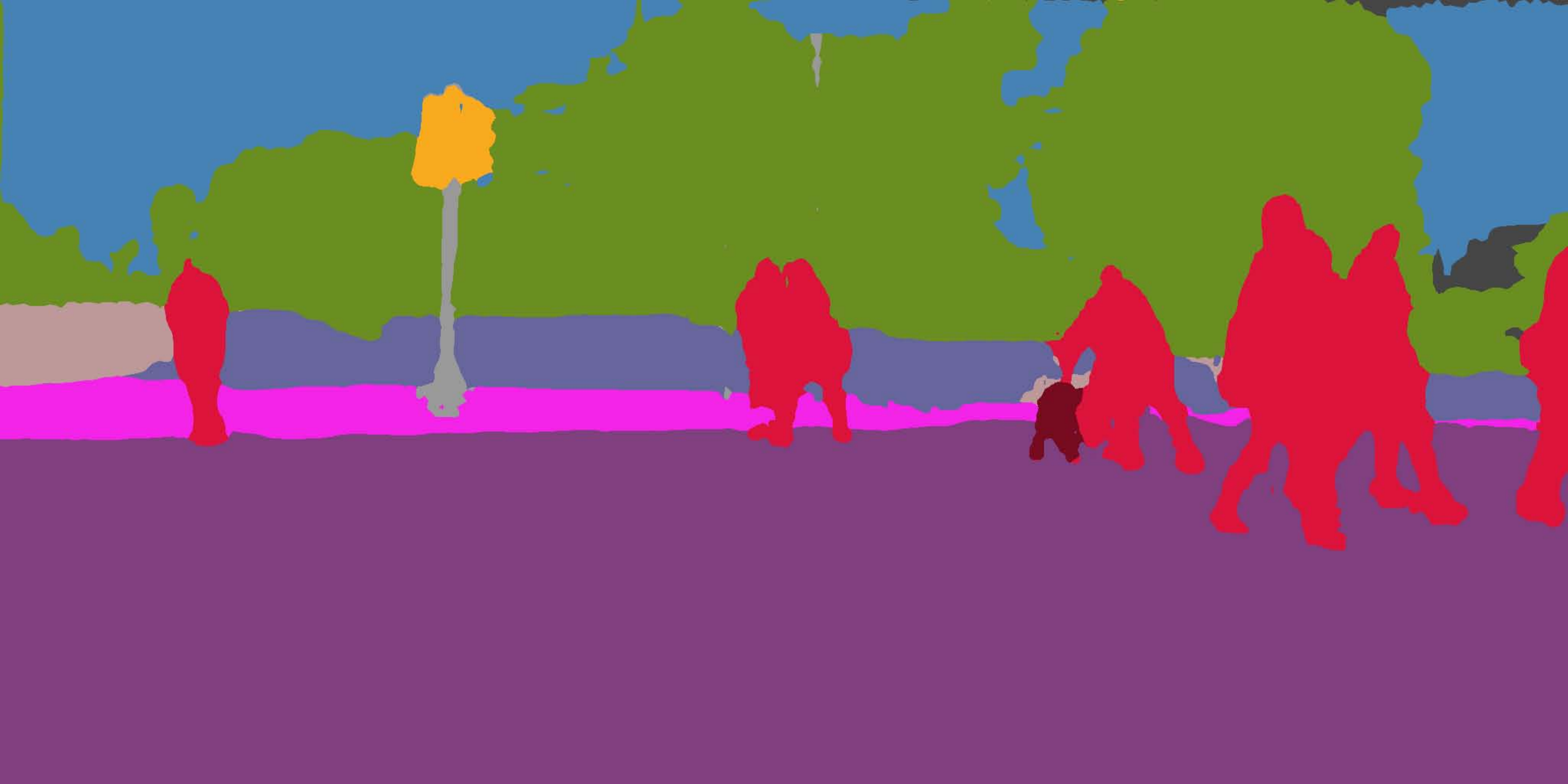}
		\includegraphics[width=0.245\textwidth, height=0.125\textwidth]{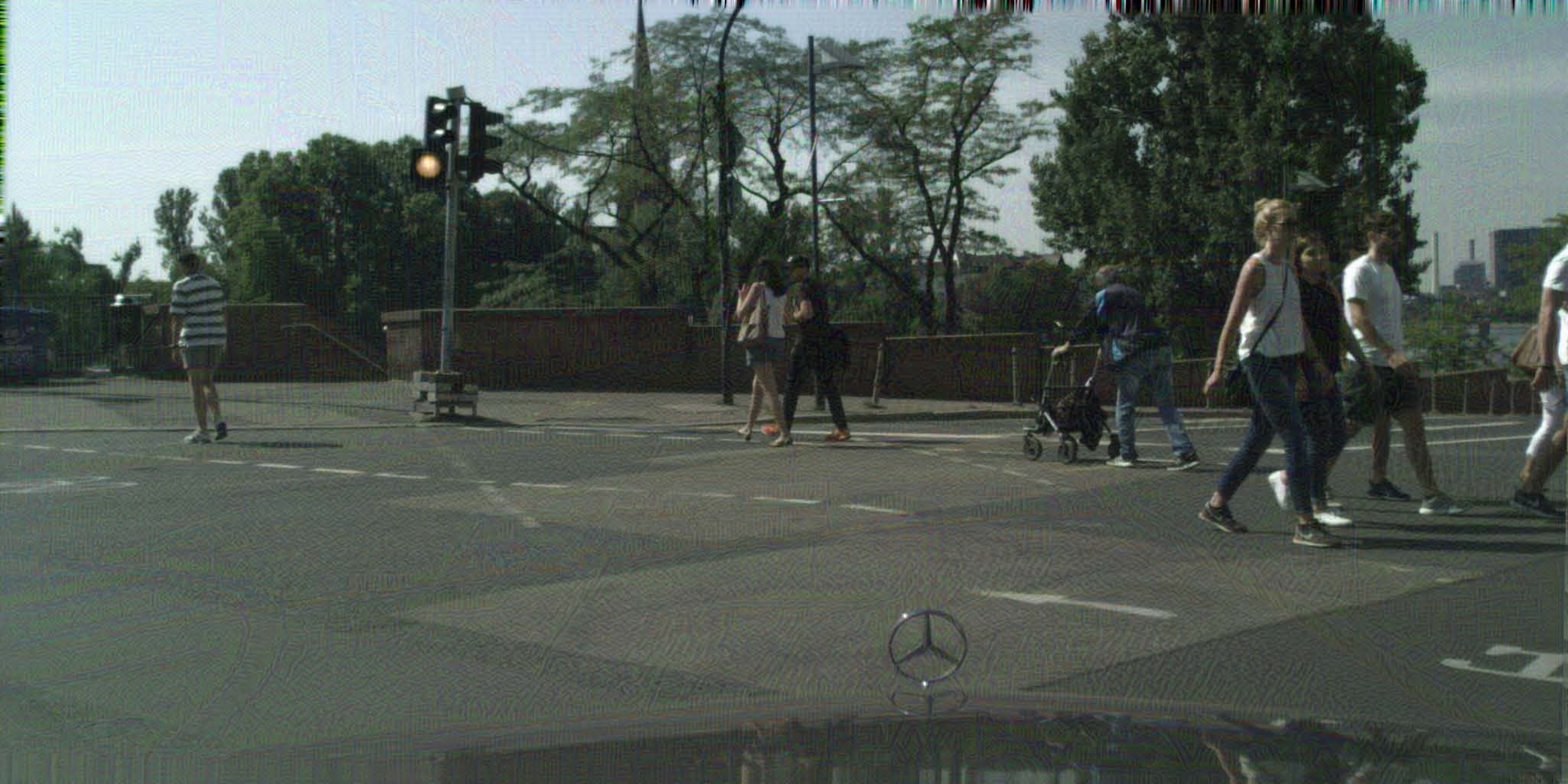}
		\includegraphics[width=0.245\textwidth, height=0.125\textwidth]{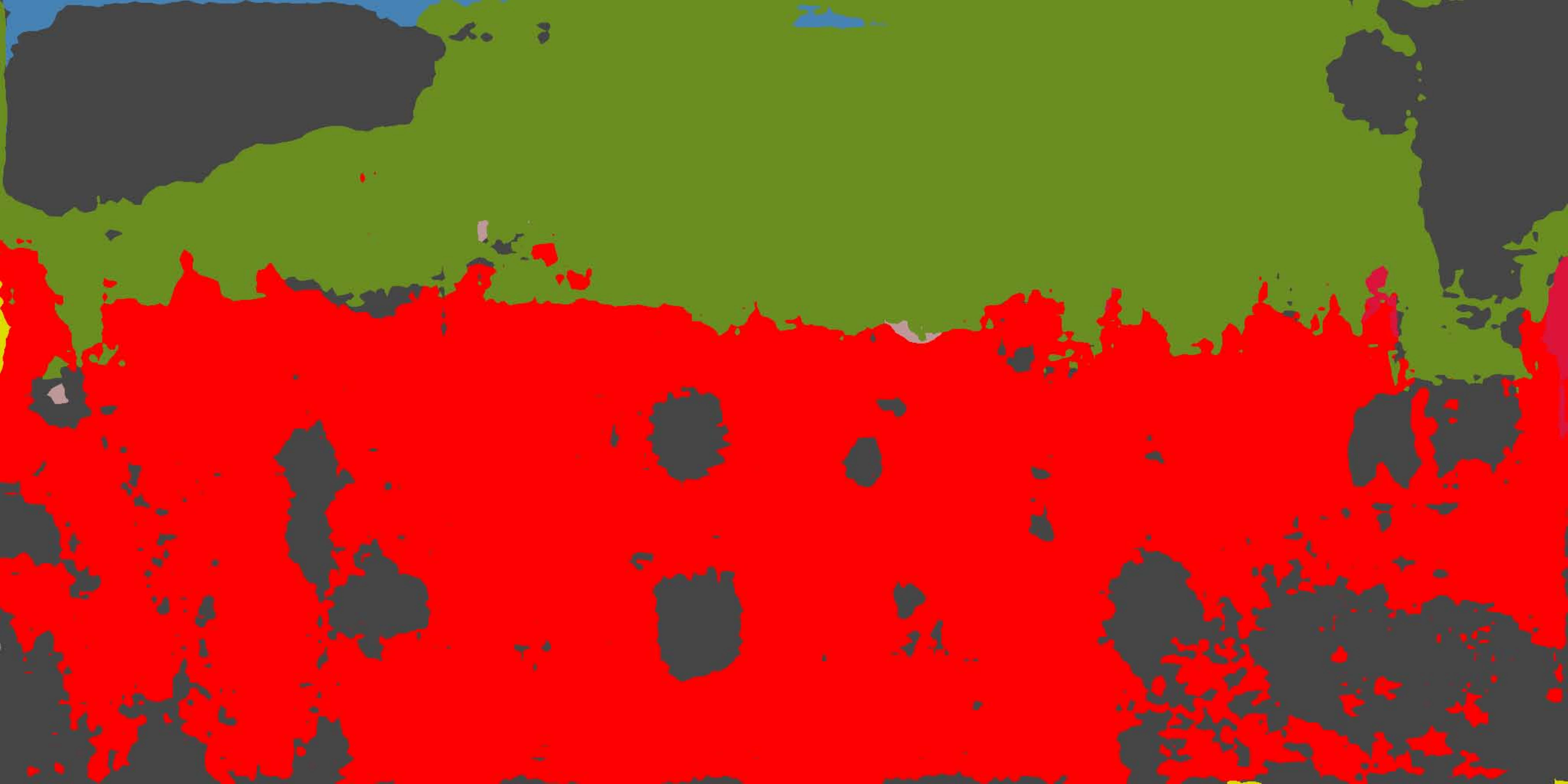}
		
		\begin{tabu} to \textwidth {X[c]X[c]X[c]X[c]}
			(a) & (b) & (c) & (d)
		\end{tabu}
		
		\caption[toc entry]{Adversarial attacks on the ICNet using a (single) universal adversarial perturbation $\boldsymbol{r}_\text{uni}$ created by Fast Feature Fool (FFF) \cite{Mopuri2018}. We show its effectiveness in fooling the ICNet on four \protect\ifthenelse{\equal{\changesred}{1}}{\red{example
			}}{example} images \protect\ifthenelse{\equal{\changesred}{1}}{\red{
					with cars or pedestrians
			}}
			{with cars or pedestrians
			}from the Cityscapes validation set. Each row corresponds to one attack scenario; (a) clean input image, (b) semantic segmentation of clean input image, (c) adversarial example created by FFF, and (d) semantic segmentation of adversarial example created by FFF.}\vspace{\myvspace}
		\label{fig:adv_mopuri_attack}
\end{figure*}}

While Moosavi-Dezfooli et al.~use samples from a set of images $\mathcal{T}$ to craft UAPs, Mopuri et al.~\cite{Mopuri2017} introduced a dataset-independent method named Fast Feature Fool (FFF).
In the following we are considering the formulation of FFF from their extended work \cite{Mopuri2018}.
Adopting the overall objective in (\ref{eq:moos_uni}), FFF aims at finding a UAP that increases the mean activation in each layer $\ell\in\mathcal{L}$, without any knowledge about the respective images $\boldsymbol{x}\in\mathcal{T}\mystrut^{\prime}$ to fool.
This is done by minimizing the following loss function
\begin{equation}
J\!\left( \boldsymbol{r}_\text{uni} \right) = -\text{log}\!\left( \prod_{\ell\in\mathcal{L}} || \boldsymbol{f}_\ell\!\left( \boldsymbol{r}_\text{uni} \right) ||_2 \right),
\end{equation}
with respect to $\boldsymbol{r}_\text{uni}$ (as 1st layer input image), constrained by (\ref{eq:moos_uni_2}) with $p=\infty$.
We used FFF on the ICNet to show the effectiveness and transferability of UAPs on several images taken from the Cityscapes validation set by following Mopuri et al.~in choosing $\epsilon=10$.
The obtained results for some images are illustrated in Fig. \ref{fig:adv_mopuri_attack}.
While not generating realistically looking semantic segmentation masks as DNNM does, FFF still completely fools the ICNet on several diverse images and needs to be computed only once to obtain $\boldsymbol{r}_\text{uni}$.
Moreover, safety-critical classes such as pedestrians and cars are removed from the scene in all examples, underlining again the risk of adversarial attacks for AD.
Note that the particular danger of this method for AD lies in the fact that it just requires a generic adversarial pattern to be added to any unknown sensorial data $\left(\boldsymbol{x} + \boldsymbol{r}_\text{uni} \right)$ during driving, causing major errors in the output segmentation mask.

\section{Adversarial Defense}
So far, we demonstrated that DNNs can be fooled in many different ways by means of almost imperceptible modifications of the input image.
This behavior of DNNs puts challenges to their application within environment perception in AD.
Therefore, appropriate adversarial defense strategies are needed to decrease the risk of DNNs being completely fooled by adversarial examples.
In this section, some adversarial defense strategies are presented, that have been hypothesized and developed to defend against adversarial attacks.
In general, adversarial defense strategies can be distinguished as being specific or agnostic to a model at hand.
In the following, we will provide a brief introduction to model-specific defense techniques, but then we will focus on model-agnostic ones.

\subsection{Model-Specific Defense Techniques}
Model-specific defense techniques aim at modifying the behavior of \ifthenelse{\equal{\changesred}{1}}{\red{a specific}}{a specific} DNN in a way that the \ifthenelse{\equal{\changesred}{1}}{\red{respective}}{respective} DNN becomes more robust towards adversarial examples.
\ifthenelse{\equal{\changesred}{1}}{\red{
		Note that such a technique most often can alternatively be applied to numerous DNN topologies, however, once being applied it always defends only the specific DNN at hand.
}}
{Note that such a technique most often can alternatively be applied to numerous DNN topologies, however, once being applied it always defends only the specific DNN at hand.
}
One well-known and intuitive method of model-specific defense techniques is \textit{adversarial training}.
In adversarial training the original training samples of the DNN are extended with their adversarial counterparts, e.g., created by FGSM from Goodfellow et al.~as shown before, and then retrained with this set of clean and adversarially perturbed images.
Whereas the performance of the DNN on adversarial examples increases (\cite{Goodfellow2015, Moosavi-Dezfooli2017}), the effect is still marginal \cite{Moosavi-Dezfooli2016}.
More importantly, it is also not clear which amount or type of adversarial examples is sufficient to increase the DNN's robustness up to a desired level.
Xie et al.~\cite{Xie2019} investigated the effect of adversarial examples on the feature maps in several layers.
Their observation was that adversarial examples create noise-like patterns in the feature maps.
To counter this, they proposed to add trainable denoising layers containing a denoising operation followed by a convolution operation.
Xie et al.~obtained the best results by using the non-local means algorithm (NLM)~\cite{Buades2005} for feature denoising.
Bär et al.~\cite{Baer2019} explored the effectiveness of teacher-student approaches in defending against adversarial attacks.
Here, an additional student DNN is included to increase the robustness against adversarial attacks, assuming that the potential attacker has a hard time to deal with a constantly adapting student DNN.
It was concluded that in combination with simple output voting schemes this approach could be a promising model-specific defense technique.
Nevertheless, a major drawback of model-specific defense techniques is that the respective DNN has to be retrained, or one has to modify the network architecture, which is not always possible when using pre-trained DNNs.

\subsection{Model-Agnostic Defense Techniques}
In contrast to model-specific defense techniques, model-agnostic \protect\ifthenelse{\equal{\changesred}{1}}{\red{
defense techniques, once developed, can be applied in conjunction with any model,
}}
{defense techniques, once developed, can be applied in conjunction with any model,
}as they do not modify the model itself but rather the input data.
\protect\ifthenelse{\equal{\changesred}{1}}{\red{
In particular, the model does not need to be retrained.
}}
{In particular, the model does not need to be retrained.
}
Hence, it serves as an image pre-processing, where the adversary is removed from the input image.

\ifthenelse{\equal{\mode}{0}}{}{
	\begin{figure*}[t!]
		\begin{tabu} to \textwidth {X[c]X[c]X[c]X[c]X[c]}
			{\fontsize{10}{12}\selectfont Clean output} & {\fontsize{10}{12}\selectfont DNNM attack ...} & {\fontsize{10}{12}\selectfont ... defended by NLM} & {\fontsize{10}{12}\selectfont ... by IQ} & {\fontsize{10}{12}\selectfont ... by NLM+IQ}
		\end{tabu}
		\includegraphics[width=\textwidth]{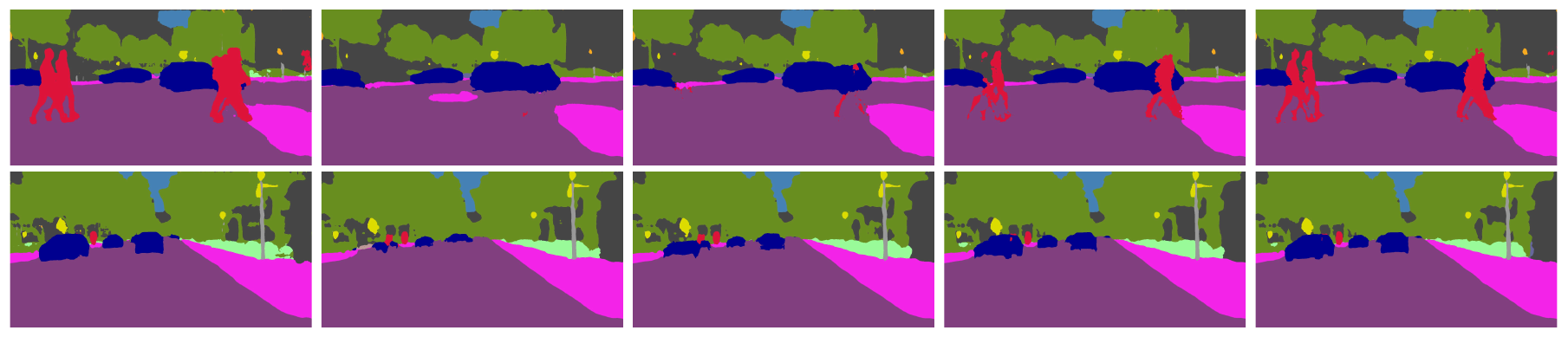}\vspace{0.1cm}
		
		\begin{tabu} to \textwidth {X[c]X[c]X[c]X[c]X[c]}
			avg. mIoU $= 67.3$ \% & $... = 57.6$ \% & $... = 60.2$ \%  & $... = 61.2$ \%  & $... = 63.3$ \%\\
			(a) & (b) & (c) & (d) & (e)
		\end{tabu}
		
		\caption[toc entry]{Adversarial attacks on the ICNet using the dynamic nearest neighbor method (DNNM) \cite{Metzen2017}, defended by image quilting (IQ) \cite{Guo2018} and the non-local means algorithm (NLM) \cite{Buades2005}. \protect\ifthenelse{\equal{\changesred}{1}}{\red{
					Both image rows correspond to the examples shown in Fig. \ref{fig:adv_metzen_attack}.
			}}
			{Both image rows correspond to the examples shown in Fig. \ref{fig:adv_metzen_attack}.
			}The first row contains an example, where DNNM was used to remove pedestrians from the scene, while the second row contains an example, where DNNM was used to remove cars instead; (a) clean output, (b) adversarial output using DNNM, (c) adversarial output using DNNM defended by NLM, (d) adversarial output using DNNM defended by IQ, and (e) adversarial output using DNNM defended by NLM and IQ combined. \ifthenelse{\equal{\changesred}{1}}{\red{The mIoU values in the bottom line refer to the \textit{average} mIoU over the entire Cityscapes validation set.}}{The mIoU values in the bottom line refer to the \textit{average} mIoU over the entire Cityscapes validation set.}}\vspace{\myvspace}
		\label{fig:adv_metzen_defense}
\end{figure*}}

Guo et al.~\cite{Guo2018} analyzed the effectiveness of non-differentiable input transformations in destroying adversarial examples.
Non-differentiability is an important property of adversarial defense strategies considering that the majority of adversarial attacks is build on gradient-based optimization.
Guo et al.~used image quilting (IQ) amongst some other input transformation techniques and observed IQ to be an effective way of performing model-agnostic defense against several adversarial attacks.
IQ is a technique, wherein the input image $\boldsymbol{x}$ is viewed as a puzzle of small patches $\boldsymbol{x}_{\mathcal{I}_i}$, with $i$ being the position of the center pixel.
To remove potential adversaries from an image, each of its patches $\boldsymbol{x}_{\mathcal{I}_i}$, irrelevant of being adversarially perturbed or not, is replaced by a nearest neighbor patch $\hat{\boldsymbol{x}}_{\mathcal{I}_i}\in\mathcal{P}\subset\mathbb{G}^{h\times b\times C}$ to obtain a quilted image $\boldsymbol{x}^\text{IQ}$, with $\mathcal{P}$ being a large set of patches created beforehand from random samples of clean images.
\protect\ifthenelse{\equal{\changesred}{1}}{\red{
		The aim is to synthetically construct an adversary-free image having the original semantic content.
}}
{The aim is to synthetically construct an adversary-free image having the original semantic content.
}

\ifthenelse{\equal{\mode}{0}}{}{
	\begin{figure*}[t!]
		\begin{tabu} to \textwidth {X[c]X[c]X[c]X[c]X[c]}
			{\fontsize{10}{12}\selectfont Clean output} & {\fontsize{10}{12}\selectfont FFF attack ...} & {\fontsize{10}{12}\selectfont ... defended by NLM} & {\fontsize{10}{12}\selectfont ... by IQ} & {\fontsize{10}{12}\selectfont ... by NLM+IQ}
		\end{tabu}
		\includegraphics[width=\textwidth]{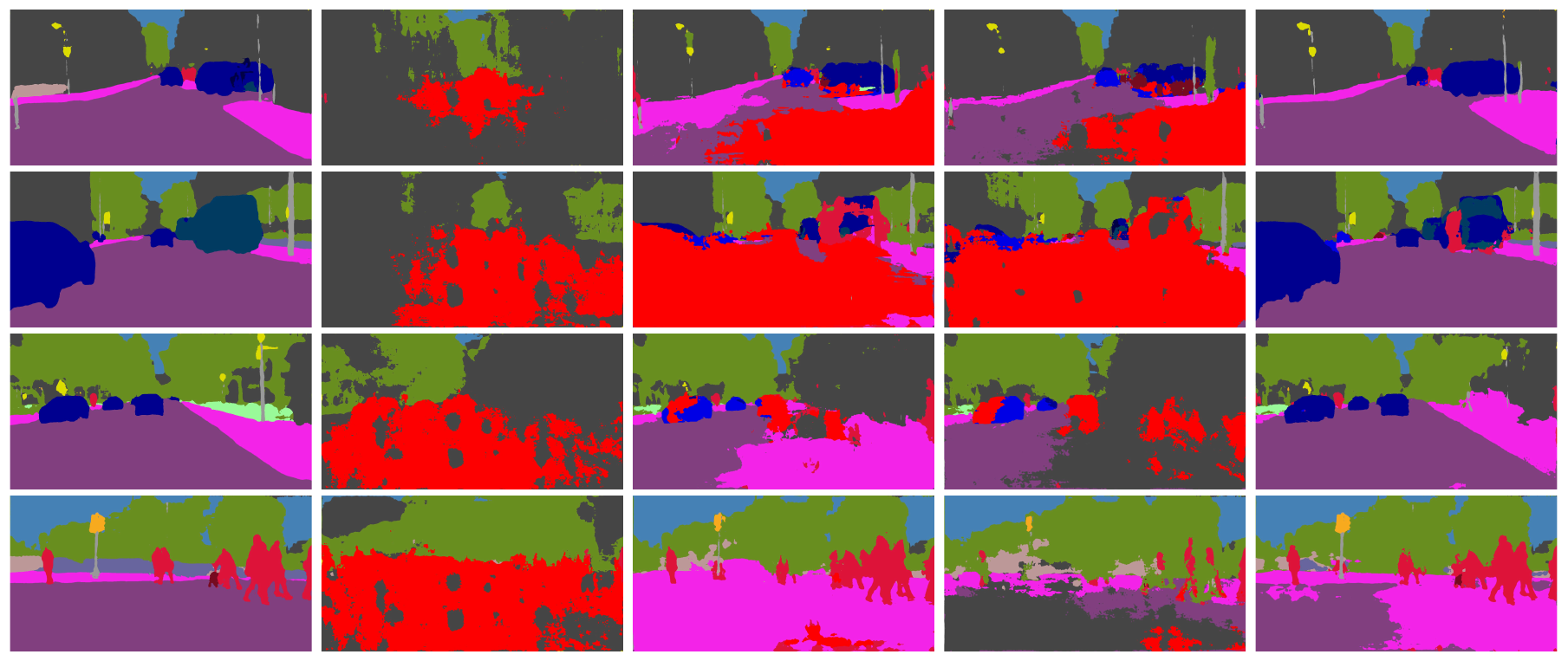}\vspace{0.1cm}
		
		\begin{tabu} to \textwidth {X[c]X[c]X[c]X[c]X[c]}
			avg. mIoU $= 67.3$ \% & $... = 4.6$ \% & $... = 25.7$ \%  & $... = 18.8$ \%  & $... = 46.0$ \%\\
			(a) & (b) & (c) & (d) & (e)
		\end{tabu}
		
		\caption[toc entry]{Adversarial attacks on the ICNet using Fast Feature Fool (FFF) \cite{Mopuri2018}, defended by image quilting (IQ) \cite{Guo2018} and the non-local means algorithm (NLM) \cite{Buades2005}. We show results on \protect\ifthenelse{\equal{\changesred}{1}}{\red{the}}{the} four semantic segmentation outputs \protect\ifthenelse{\equal{\changesred}{1}}{\red{from Fig.~\ref{fig:adv_mopuri_attack} using}}{from Fig.~\ref{fig:adv_mopuri_attack} using} the Cityscapes validation set. Each row corresponds to an attack scenario to be defended by NLM, IQ, or a combination of both; (a) clean output, (b) adversarial output using FFF, (c) adversarial output using FFF defended by NLM, (d) adversarial output using FFF defended by IQ, and (e) adversarial output using FFF defended by NLM and IQ combined. \protect\ifthenelse{\equal{\changesred}{1}}{\red{The mIoU values in the bottom line refer to the \textit{average} mIoU over the entire Cityscapes validation set.}}{The mIoU values in the bottom line refer to the \textit{average} mIoU over the entire Cityscapes validation set.}}\vspace{\myvspace}
		\label{fig:adv_mopuri_defense}
\end{figure*}}

Another model-agnostic defense technique is the non-local means algorithm (NLM) from Buades et al.~\cite{Buades2005}.
NLM aims at denoising the input image.
To accomplish this, NLM replaces each pixel value $x_i$ by
\begin{equation}\label{eq:NLM_1}
x^\text{NLM}_i =\sum_{j\in\mathcal{I}} w_{i,j}\, x_j,
\end{equation}

with the NLM-denoised pixel $x^\text{NLM}_i$, the inter-pixel weighting factor {$w_{i,j}\in [0, 1]$} for which $\sum_{j\in\mathcal{I}} w_{i,j} = 1$ holds, and the pixel value $x_j$ at position $j$.
The inter-pixel weighting factor $w_{i,j}$ \protect\ifthenelse{\equal{\changesred}{1}}{\red{relates}}{relates} the respective pixel $x_i$ at pixel position $i$ \protect\ifthenelse{\equal{\changesred}{1}}{\red{to}}{to}  the pixel $x_j$ at pixel position $j$.
It is defined by
\begin{equation}
w_{i,j} = \frac{1}{\alpha_i} \,\text{exp} \left( -\frac{||\boldsymbol{x}_{\mathcal{I}_i} - \boldsymbol{x}_{\mathcal{I}_j} ||^2_{2,a}}{h^2}\right),
\end{equation}
with the patches $\boldsymbol{x}_{\mathcal{I}_i}$ and $\boldsymbol{x}_{\mathcal{I}_j}$ centered at pixel positions $i$ and $j$, the squared Gaussian weighted Euclidean distance $||\cdot ||^2_{2, a}$, with $a>0$ as the standard deviation of the Gaussian kernel, the hyperparameter for the degree of filtering $h$, and the normalizing factor $\alpha_i$.
By incorporating the squared Gaussian weighted Euclidean distance, a large weight is put to pixels $x_j$, whose neighborhood $\boldsymbol{x}_{\mathcal{I}_j}$ looks similar to $\boldsymbol{x}_{\mathcal{I}_i}$ (the neighborhood of the respective pixel $x_i$ to be denoised).

The idea behind NLM is to remove the high local dependency of adversarial perturbations.
Nevertheless, applying NLM on the complete input image, as stated in (\ref{eq:NLM_1}), can be computationally demanding.
Thus, the search window is often reduced to an image region $\mathcal{R}_i\subset\mathcal{I}$ of size $|\mathcal{R}_i|=R\times R$.
Note that $\mathcal{I}_i\subset\mathcal{R}_i$, with $|\mathcal{I}_i|=|\mathcal{I}_j|<|\mathcal{R}_i|$.

Now let's look into results of both model-agnostic defense methods, IQ and NLM, on the adversarial examples shown before.
For IQ, the patch dataset $\mathcal{P}$ was created using samples from the Cityscapes training set.
Here, we followed Guo et al.~and collected $|\mathcal{P}|=1,000,000$ patches of size $5\times 5$ pixels in total.
Increasing the size of the patch dataset will lead to better approximations of the patches, but on the other hand also increases the search space.
The same holds when decreasing the size of the patches up to a certain level.

\ifthenelse{\equal{\mode}{0}}{
	\begin{figure*}[t!]
		\begin{tabu} to \textwidth {X[c]X[c]X[c]X[c]X[c]}
			{\fontsize{9}{12}\selectfont Clean output} & {\fontsize{9}{12}\selectfont DNNM attack ...} & {\fontsize{9}{12}\selectfont\mbox{... defended by NLM}} & {\fontsize{9}{12}\selectfont ... by IQ} & {\fontsize{9}{12}\selectfont ... by NLM+IQ}
		\end{tabu}
		
		\includegraphics[width=\textwidth]{./fig/fig_metzen_def.pdf}\vspace{0.1cm}

		\begin{tabu} to \textwidth {X[c]X[c]X[c]X[c]X[c]}
			\mbox{\fontsize{9}{1}\selectfont avg.~mIoU $= 67.3$ \% }& {\fontsize{9}{12}\selectfont $... = 57.6$ \%} & {\fontsize{9}{12}\selectfont $... = 60.2$ \% } & {\fontsize{9}{12}\selectfont $... = 61.2$ \% } & {\fontsize{9}{12}\selectfont $... = 63.3$ \%}\\
			(a) & (b) & (c) & (d) & (e)
		\end{tabu}
		
		\caption[toc entry]{Adversarial attacks on the ICNet using the dynamic nearest neighbor method (DNNM) \cite{Metzen2017}, defended by image quilting (IQ) \cite{Guo2018} and the non-local means algorithm (NLM) \cite{Buades2005}. 		\protect\ifthenelse{\equal{\changesred}{1}}{\red{
					Both image rows correspond to the examples shown in Fig. \ref{fig:adv_metzen_attack}.
			}}
			{Both image rows correspond to the examples shown in Fig. \ref{fig:adv_metzen_attack}.
			}The first row contains an example, where DNNM was used to remove pedestrians from the scene, while the second row contains an example, where DNNM was used to remove cars instead; (a) clean output, (b) adversarial output using DNNM, (c) adversarial output using DNNM defended by NLM, (d) adversarial output using DNNM defended by IQ, and (e) adversarial output using DNNM defended by NLM and IQ combined. \protect\ifthenelse{\equal{\changesred}{1}}{\red{The mIoU values in the bottom line refer to the \textit{average} mIoU over the entire Cityscapes validation set.}}{The mIoU values in the bottom line refer to the \textit{average} mIoU over the entire Cityscapes validation set.}}\vspace{\myvspace}
		\label{fig:adv_metzen_defense}
\end{figure*}}{}

For NLM, patches $\mathcal{I}_i$ and $\mathcal{I}_j$ of size $7\times 7$ were used and the image region for neighbor pixel was restricted according to $|\mathcal{R}_i|=9\times9$ to keep an adequate algorithm complexity.
The degree of filtering $h$ was computed by $h=2.15\,\tilde{\sigma}\!\left( \boldsymbol{x}\right)$,
with $\tilde{\sigma}\!\left( \boldsymbol{x}\right)$ being an estimate for the Gaussian noise standard deviation on the input image $\boldsymbol{x}$.

Using these settings, we tested IQ, NLM, as well as a combined version of both, denoted as \text{IQ+NLM}, on the adversarial attacks shown in Section \ref{sec:adversarial_attacks} (see Fig.~\ref{fig:adv_metzen_attack} and Fig.~\ref{fig:adv_mopuri_attack}).
It is important to note that we applied both defense methods without any extensive hyperparameter search. 
The adversarial defenses on \textit{DNNM-attacked images} are depicted in Fig.~\ref{fig:adv_metzen_defense}.
From left to right, the original semantic segmentation mask is reconstructed better and better, with the combination of NLM and IQ showing the best results (Fig.~7 (e)).
Comparing NLM and IQ separately, it can be seen that IQ is able to reconstruct the original semantic segmentation mask even more precisely.
\ifthenelse{\equal{\changesred}{1}}{\red{The same behavior can be observed when looking at the mIoU values in Fig.~\ref{fig:adv_metzen_defense}, where we report averages over the entire Cityscapes validation set.}}{The same behavior can be observed when looking at the mIoU values in Fig.~\ref{fig:adv_metzen_defense} where we report averages over the entire Cityscapes validation set.}
\ifthenelse{\equal{\changesred}{1}}{\red{Altogether, the results show that by combining NLM with IQ one can lever the destructiveness of DNNM---an important and releaving observation.}}{Altogether, the results show that by combining NLM with IQ one can lever the destructiveness of DNNM---an important and releaving observation.}

The adversarial defenses on \textit{FFF-attacked images} are illustrated \ifthenelse{\equal{\changesred}{1}}{\red{and supported by the corresponding \textit{average} mIoU values on the Cityscapes validation set in Fig.~\ref{fig:adv_mopuri_defense}}}{and supported by the corresponding \textit{average} mIoU values on the Cityscapes validation set in Fig.~\ref{fig:adv_mopuri_defense}}.
\ifthenelse{\equal{\changesred}{1}}{\red{Here,}}{Here,} it is not trivial to judge by only looking at the images, which defense is superior\ifthenelse{\equal{\changesred}{1}}{\red{, IQ or NLM}}{, IQ or NLM}.
In some cases, NLM \ifthenelse{\equal{\changesred}{1}}{\red{seems to lead}}{seems to lead} to the better results, whereas in other cases IQ \ifthenelse{\equal{\changesred}{1}}{\red{seems to outperform}}{seems to outperform} NLM.
\ifthenelse{\equal{\changesred}{1}}{\red{Yet, looking at the average mIoU values for the entire Cityscapes validation set leads to the conclusion that overall NLM is superior to IQ.}}{Yet, looking at the average mIoU values for the entire Cityscapes validation set leads to the conclusion that overall NLM is superior to IQ.}
\ifthenelse{\equal{\changesred}{1}}{\red{Moreover, combining NLM with IQ again shows the best results leading to an overall significant improvement in restoration of the segmentation masks.}}{Moreover, combining NLM with IQ again shows the best results leading to an overall significant improvement in restoration of the segmentation masks.}
This observation is both extremely important and relieving, as the existence of UAPs is particularly dangerous for the use case of DNNs in AD.

\ifthenelse{\equal{\changesred}{1}}{\red{
		Even though we observe a certain level of effectiveness in using model-agnostic defense methods, there is still room left for improvement in defending against adversarial attacks.
		The work of Carlini and Wagner \cite{Carlini2017} and Athalye et al.~\cite{Athalye2018} are just two of many representative examples.
		Carlini and Wagner bypassed several state-of-the-art detection systems for adversarial examples with their approach, whereas Athalye et al.~circumvented the non-differentiality property of some state-of-the-art defenses by different gradient approximation methods.
}}{Even though we observe a certain level of effectiveness in using model-agnostic defense methods, there is still room left for improvement in defending against adversarial attacks.
	The work of Carlini and Wagner \cite{Carlini2017} and Athalye et al.~\cite{Athalye2018} are just two of many representative examples.
	Carlini and Wagner bypassed several state-of-the-art detection systems for adversarial examples with their approach, whereas Athalye et al.~circumvented the non-differentiality property of some state-of-the-art defenses by different gradient approximation methods.
}

\section{Summary and Future Directions}
Deep neural networks (DNNs) are one of the most promising technologies for the use case of environment perception in autonomous driving (AD).
Assuming the environment perception system consists of several camera sensors, a DNN trained for semantic segmentation can be used to perform extensive environment sensing in real-time.
Nevertheless, today's state-of-the-art DNNs still unveil flaws when fed with specifically crafted inputs, denoted as adversarial examples.
It was step-by-step demonstrated that it is quite easy and intuitive to craft adversarial examples for individual input images using the least-likely class method (LLCM) or the dynamic nearest neighbor method (DNNM) by simply performing gradient updates on the clean input image.
It is even possible to craft adversarial examples to fool not only one but a set of images using the Fast Feature Fool (FFF) method, without any knowledge of the respective input image to be perturbed.
This in turn highlights the importance of appropriate defense strategies.
From a safety-concerned perspective, the lack of robustness shown by DNNs is a highly relevant and important challenge to deal with, before AD vehicles are released for public use.

DNNs' lack of robustness evoked the need for defense strategies and other fallback strategies regarding the safety relevance for AD applications.
Model-agnostic defense strategies only modify the potentially perturbed input image to decrease the effect of adversarial attacks.
This way, an already pretrained DNN can be used without the need of retraining or modifying the DNN itself.
We explored two model-agnostic defense strategies, namely image quilting (IQ) and the non-local means algorithm (NLM), both on DNNM and FFF attacks, where the combination of IQ and NLM shows the best results on almost all images.
Nevertheless, although clearly robustifying the DNNs towards adversarial attacks, the current state of research in model-agnostic defense strategies also showed that \ifthenelse{\equal{\changesred}{1}}{\red{vulnerability of DNNs is not entirely}}{vulnerability of DNNs is not entirely} solved yet.
However, \ifthenelse{\equal{\changesred}{1}}{\red{ensembles}}{ensembles} of model-agnostic defenses could be promising \ifthenelse{\equal{\changesred}{1}}{\red{for}}{for} tackling adversarial attacks, as well as intelligent redundancy, e.g., by teacher-student approaches.
\ifthenelse{\equal{\changesred}{1}}{\red{We would also like to point out that certification methods (\cite{Dvijotham2018, Wu2018}) should be further investigated to really obtain provable robustness.}}{We would also like to point out that certification methods (\cite{Dvijotham2018, Wu2018}) should be further investigated to really obtain provable robustness.}

What does this mean regarding the application of DNNs for AD?
Are today's DNNs not suitable for safety-critical applications in AD?
We would argue that this is to some extent true, if we only consider applying model-agnostic defenses \ifthenelse{\equal{\changesred}{1}}{\red{without certification}}{without certification}.
DNN training and DNN understandability are two highly dynamic academic fields of research.
Research so far mainly focused on increasing the performance of DNNs, widely neglecting their robustness \ifthenelse{\equal{\changesred}{1}}{\red{and certification}}{and certification}.
In order to develop \ifthenelse{\equal{\changesred}{1}}{\red{employable}}{employable} machine learning-based functions that are realistically usable in a real world setting, it is extremely important to establish their robustness against slight input alterations in addition to improving the task performance.
Furthermore, new mature defense \ifthenelse{\equal{\changesred}{1}}{\red{and certification}}{and certification} strategies are needed, including fusion approaches, redundancy concepts, and modern fallback strategies.
\ifthenelse{\equal{\changesred}{1}}{\red{We especially recommend automotive companies to focus on certication of DNNs.}}{We especially recommend automotive companies to focus on certication of DNNs.}
Otherwise, doors would open for potentially fatal attacks which in turn would have consequences on public acceptance of AD.

\section*{Acknowledgement}
The authors gratefully acknowledge support of this work by Volkswagen Group Automation, Wolfsburg, Germany\ifthenelse{\equal{\changesred}{1}}{\red{
		, and would like to thank Nico M. Schmidt and Zeyun Zhong for their help in setting up final experiments.
}}
{, and would like to thank Nico M. Schmidt and Zeyun Zhong for their help in setting up final experiments.}

\section*{Authors}
\textbf{Andreas Bär} (andreas.baer@tu-bs.de) received his B.Eng. degree from Ostfalia University of Applied Sciences, Wolfenbüttel, Germany, in 2016, and his M.Sc. degree from Technische Universität Braunschweig, Braunschweig, Germany, in 2018, where he is currently a Ph.D. degree candidate in the Faculty of Electrical Engineering, Information Technology, and Physics. 
His research interests include convolutional neural networks for camera-based environment perception and the robustness of neural networks to adversarial attacks.
In 2020, he won the Best Paper Award at the Workshop on Safe Artificial Intelligence for Automated Driving, held in conjunction with the IEEE Conference on Computer Vision and Pattern Recognition, along with coauthors Serin John Varghese, Fabian Hüger, Peter Schlicht, and Tim Fingscheidt.

\textbf{Jonas Löhdefink} (j.loehdefink@tu-bs.de) received his B. Eng. degree from Ostfalia University of Applied Sciences, Wolfenbüttel, Germany, in 2015, and his M.Sc. degree from Technische Universität Braunschweig, Braunschweig, Germany, in 2018, where he is currently a Ph.D. degree candidate in the Faculty of Electrical Engineering, Information Technology, and Physics.
His research interests include learned image compression and quantization approaches by means of convolutional neural networks and generative adversarial networks.

\textbf{Nikhil Kapoor} (nikhil.kapoor@volkswagen.de) reeived his B.Eng. degree from the Army Institute of Technology, Pune, India, in 2012, and his M.Sc. degree from RWTH Aachen University, Germany, in 2018.
Currently, he is a Ph.D. degree candidate at Technische Universität Braunschweig, Braunschweig, Germany, in cooperation with Volkswagen Group Research.
His research focuses on training strategies that range from improving the robustness of neural networks for camera-based perception tasks to augmentations and adversarial perturbations using concept-based learning.

\textbf{Serin John Varghese} (john.serin.varghese@volkswagen.de) received his B.Eng. degree from the University of Pune, India, in 2013, and his M.Sc. degree from Technische Universität Chemnitz, Germany, in 2018.
Currently, he is a Ph.D. degree candidate at Technische Universität Braunschweig, Braunschweig, Germany, in cooperation with Volkswagen Group Research.
His research is focused on compression techniques for convolutional neural networks used for perception modules in automated driving, with a focus on not only inference times but also maintaining, and even improving, the robustness of neural networks.

\textbf{Fabian Hüger} (fabian.hueger@volkswagen.de) received his M.Sc. degree in electrical and computer engineering from the University of California, Santa Barbara, as a Fulbright scholar in 2009.
He received his Dipl.-Ing. and Dr.-Ing. degrees in electrical engineering from the University of Kassel, Germany, in 2010 and 2014, respectively.
He joined Volkswagen Group Research, Germany, in 2010, and his current research is focused on safe and efficient use of artificial intelligence for autonomous driving.

\textbf{Peter Schlicht} (peter.schlicht@volkswagen.de) received his Ph.D. degree in mathematics from the University of Leipzig, Germany.
After a two-year research stay at the Ecole Polytechnique Fédérale, Lausanne, Switzerland, he joined Volkswagen Group Research, Wolfsburg, Germany, in 2016 as an artificial intelligence (AI) architect.
There he deals with research questions on AI technologies for automatic driving.
His research interests include methods used for monitoring, explaining, and robotizing deep neural networks as well as securing them.

\textbf{Tim Fingscheidt} (t.fingscheidt@tu-bs.de) received his Dipl.-Ing. and Ph.D. degrees in electrical engineering, both from RWTH Aachen University, Germany, in 1993 and 1998, respectively.
Since 2006, he has been a full professor with the Institute for Communications Technology, Technische Universität Braunschweig, Braunschweig, Germany.
He received the Vodafone Mobile Communications Foundation prize in 1999 and the 2002 prize of the Information Technology branch of the Association of German Electrical Engineers (VDE ITG).
In 2017, he coauthored the ITG award-winning publication, “Turbo Automatic Speech Recognition.”
He has been the speaker of the Speech Acoustics Committee ITG AT3 since 2015.
He served as an associate editor of IEEE Transactions on Audio, Speech, and Language Processing (2008–2010) and was a member of the IEEE Speech and Language Processing Technical Committee (2011–2018).
His research interests include speech technology and vision for autonomous driving. He is a Senior Member of IEEE.

{\small
	\bibliographystyle{ieee}
	\bibliography{manuscript}
}

\end{document}

%% file: fig_source/tu_bs_colors.tex
\definecolor{tu0}{rgb}{0.7451, 0.1176, 0.2353}

\definecolor{tu1}{rgb}{1.0000, 0.8039, 0.0000}
\definecolor{tu11}{rgb}{1.0000, 0.8627, 0.3020}
\definecolor{tu12}{rgb}{1.0000, 0.9020, 0.4980}
\definecolor{tu13}{rgb}{1.0000, 0.9412, 0.6980}
\definecolor{tu14}{rgb}{1.0000, 0.9608, 0.8000}

\definecolor{tu2}{rgb}{0.9804, 0.4314, 0.0000}
\definecolor{tu21}{rgb}{0.9882, 0.6039, 0.3020}
\definecolor{tu22}{rgb}{0.9882, 0.7137, 0.4980}
\definecolor{tu23}{rgb}{0.9922, 0.8275, 0.6980}
\definecolor{tu24}{rgb}{0.9961, 0.8863, 0.8000}

\definecolor{tu3}{rgb}{0.6902, 0.0000, 0.2745}
\definecolor{tu31}{rgb}{0.7529, 0.2000, 0.4196}
\definecolor{tu32}{rgb}{0.8431, 0.4980, 0.6353}
\definecolor{tu33}{rgb}{0.9216, 0.7490, 0.8196}
\definecolor{tu34}{rgb}{0.9529, 0.8510, 0.8902}

\definecolor{tu4}{rgb}{0.4863, 0.8039, 0.9020}
\definecolor{tu41}{rgb}{0.6431, 0.8627, 0.9333}
\definecolor{tu42}{rgb}{0.7412, 0.9020, 0.9490}
\definecolor{tu43}{rgb}{0.8431, 0.9412, 0.9686}
\definecolor{tu44}{rgb}{0.8980, 0.9608, 0.9804}

\definecolor{tu5}{rgb}{0.0000, 0.5020, 0.7059}
\definecolor{tu51}{rgb}{0.3020, 0.6510, 0.7961}
\definecolor{tu52}{rgb}{0.5490, 0.7765, 0.8667}
\definecolor{tu53}{rgb}{0.7490, 0.8745, 0.9255}
\definecolor{tu54}{rgb}{0.8510, 0.9255, 0.9569}

\definecolor{tu6}{rgb}{0.0000, 0.3255, 0.4549}
\definecolor{tu61}{rgb}{0.2510, 0.4941, 0.5922}
\definecolor{tu62}{rgb}{0.5490, 0.6941, 0.7529}
\definecolor{tu63}{rgb}{0.7490, 0.8314, 0.8627}
\definecolor{tu64}{rgb}{0.8510, 0.8980, 0.9176}

\definecolor{tu7}{rgb}{0.7765, 0.9333, 0.0000}
\definecolor{tu71}{rgb}{0.8431, 0.9529, 0.3020}
\definecolor{tu72}{rgb}{0.8863, 0.9647, 0.4980}
\definecolor{tu73}{rgb}{0.9333, 0.9804, 0.6980}
\definecolor{tu74}{rgb}{0.9569, 0.9882, 0.8000}

\definecolor{tu8}{rgb}{0.5373, 0.6431, 0.0000}
\definecolor{tu81}{rgb}{0.6784, 0.7490, 0.3020}
\definecolor{tu82}{rgb}{0.7686, 0.8196, 0.4980}
\definecolor{tu83}{rgb}{0.8588, 0.8941, 0.6980}
\definecolor{tu84}{rgb}{0.9059, 0.9294, 0.8000}

\definecolor{tu9}{rgb}{0.0000, 0.4431, 0.3373}
\definecolor{tu91}{rgb}{0.3020, 0.6118, 0.5373}
\definecolor{tu92}{rgb}{0.5490, 0.7490, 0.7020}
\definecolor{tu93}{rgb}{0.7490, 0.8588, 0.8353}
\definecolor{tu94}{rgb}{0.8549, 0.9176, 0.9059}

\definecolor{tu10}{rgb}{0.8000, 0.0000, 0.6000}
\definecolor{tu101}{rgb}{0.8706, 0.3490, 0.7412}
\definecolor{tu102}{rgb}{0.9216, 0.6000, 0.8392}
\definecolor{tu103}{rgb}{0.9608, 0.8000, 0.9216}
\definecolor{tu104}{rgb}{0.9804, 0.8980, 0.9608}

\definecolor{tu110}{rgb}{0.4627, 0.0000, 0.4627}
\definecolor{tu111}{rgb}{0.5961, 0.2510, 0.5961}
\definecolor{tu112}{rgb}{0.7294, 0.4980, 0.7294}
\definecolor{tu113}{rgb}{0.8392, 0.6980, 0.8392}
\definecolor{tu114}{rgb}{0.9216, 0.8510, 0.9216}

\definecolor{tu120}{rgb}{0.4627, 0.0000, 0.3294}
\definecolor{tu121}{rgb}{0.6118, 0.3020, 0.5333}
\definecolor{tu122}{rgb}{0.7569, 0.5490, 0.6980}
\definecolor{tu123}{rgb}{0.8667, 0.7490, 0.8314}
\definecolor{tu124}{rgb}{0.9216, 0.8510, 0.9020}

\definecolor{tu130}{rgb}{0.0314, 0.0314, 0.0314}
\definecolor{tu131}{rgb}{0.3725, 0.3725, 0.3725}
\definecolor{tu132}{rgb}{0.5882, 0.5882, 0.5882}
\definecolor{tu133}{rgb}{0.7529, 0.7529, 0.7529}
\definecolor{tu134}{rgb}{0.8667, 0.8667, 0.8667}